\documentclass{article} % For LaTeX2e
\usepackage{colm2024_conference}

\usepackage{microtype}
\usepackage{hyperref}
\usepackage{url}
\usepackage{booktabs}
\usepackage[]{mdframed}
\usepackage{framed}

\usepackage{amssymb}% http://ctan.org/pkg/amssymb
\usepackage{pifont}% http://ctan.org/pkg/pifont

\definecolor{darkblue}{rgb}{0, 0, 0.5}
\hypersetup{colorlinks=true, citecolor=darkblue, linkcolor=darkblue, urlcolor=darkblue}

\newcommand{\pro}{\texttt{Gemini-Pro}\xspace}
\newcommand{\nano}{\texttt{Gemini-Nano}\xspace}
\newcommand{\chatgpt}{\texttt{GPT3.5}\xspace}
\newcommand{\update}{UPDATE\xspace}
\newcommand{\merge}{MERGE\xspace}

\newcommand\blfootnote[1]{%
  \begingroup
  \renewcommand\thefootnote{}\footnote{#1}%
  \addtocounter{footnote}{-1}%
  \endgroup
}

\title{\datasetname: A Synthetic Benchmark for Incremental Entity Summarization}

\author{Eunjeong Hwang$^*$$^{\dagger}$\\
University of British Columbia\\
Vancouver, CA \\
\texttt{ejhwang@cs.ubc.ca} \\
\AND
Yichao Zhou$^*$, Beliz Gunel, James Bradley Wendt \& Sandeep Tata \\
Google Deepmind \\
Mountain View, USA \\
\texttt{\{yichaojoey, bgunel, jwendt, tata\}@google.com}
}

\usepackage{xspace}
\usepackage{xcolor}
\usepackage{enumitem}
\usepackage{bm}
\usepackage{graphicx}
\usepackage{subcaption}
\usepackage{amsmath}
\usepackage{url}
\usepackage{array,multirow}
\usepackage[utf8]{inputenc}
\usepackage[english]{babel}

\usepackage{amsthm}
\theoremstyle{plain}

\theoremstyle{remark}

\usepackage{amssymb}
\usepackage{adjustbox}
\usepackage[ruled,vlined]{algorithm2e} %linesnumbered

\newcommand{\datasetname}{\texttt{SUMIE}\xspace}

\colmfinalcopy % Uncomment for camera-ready version, but NOT for submission.
\begin{document}

\maketitle
\blfootnote{$^*$Equal contribution.}
\blfootnote{$^{\dagger}$This work was completed while the author was working as an intern at Google Deepmind.}

\begin{abstract}
No existing dataset adequately tests how well language models can incrementally update entity summaries – a crucial ability as these models rapidly advance. The Incremental Entity Summarization (IES) task is vital for maintaining accurate, up-to-date knowledge.  To address this, we introduce \datasetname, a fully synthetic dataset designed to expose real-world IES challenges. This dataset effectively highlights problems like incorrect entity association and incomplete information presentation. Unlike common synthetic datasets, ours captures the complexity and nuances found in real-world data. We generate informative and diverse attributes, summaries, and unstructured paragraphs in sequence, ensuring high quality. The alignment between generated summaries and paragraphs exceeds 96\%, confirming the dataset's quality. Extensive experiments demonstrate the dataset's difficulty – state-of-the-art LLMs struggle to update summaries with an F1 higher than 80.4\%. We will open source the benchmark and the evaluation metrics to help the community make progress on IES tasks. \footnote{We will release the \datasetname dataset at \url{https://github.com/google-research-datasets/sumie} soon.}
\end{abstract}

\section{Introduction}

% What are entity summarization and incremental ...? Explain the impact.
% Entity Summarization (ES) distills the essential characteristics of an entity (a person, place, organization, product, etc.) from vast unstructured knowledge sources. This task is fundamental to many NLP applications, such as question answering systems~\citep{allam2012question}, information retrieval systems~\citep{kowalski2007information}, and pairwise entity comparison systems~\citep{gunel2023strum}, providing concise and informative snapshots of key entities. Traditional ES tasks focus on computing concise summaries for entities, drawing on a size-limited selection of triples (subject-predicate-object statements) within structured RDF data~\citep{liu2020esbm,liu2021entity}. This work goes further, creating precise and comprehensive structured summaries for entities by leveraging the vast knowledge available in natural language on the web. This addresses real-world challenges and has wide-ranging applications, particularly for emerging entities primarily documented in recent web sources. For instance, comprehensive hotel and restaurant summaries eliminate the need to gather information from multiple sources, allowing users to quickly research destinations. Additionally, structured entity summaries enable easy comparisons of detailed lodging information, empowering travelers to make informed choices that perfectly match their preferences.

Entity Summarization (ES) distills key features of entities (e.g., people, places, organizations) from extensive unstructured data, essential for various NLP applications like question answering~\citep{allam2012question}, information retrieval~\citep{kowalski2007information}, and entity comparison systems~\citep{gunel2023strum}. Traditional ES tasks focus on computing concise summaries for entities, drawing on a size-limited selection of triples (subject-predicate-object statements) within structured RDF data~\citep{liu2020esbm,liu2021entity}. This work goes further, creating precise and comprehensive structured summaries for entities by leveraging the vast knowledge available in natural language on the web. This method tackles real-world issues, enhancing the utility for new entities mainly found online. For example, it streamlines gathering comprehensive summaries for hotels and restaurants, facilitating efficient destination research. Moreover, structured summaries simplify comparing detailed lodging options, aiding travelers in making choices that align with their preferences.

\begin{figure*}[t]
\small
    \centering
    \vspace{-30pt}
    \includegraphics[width=.9\linewidth]{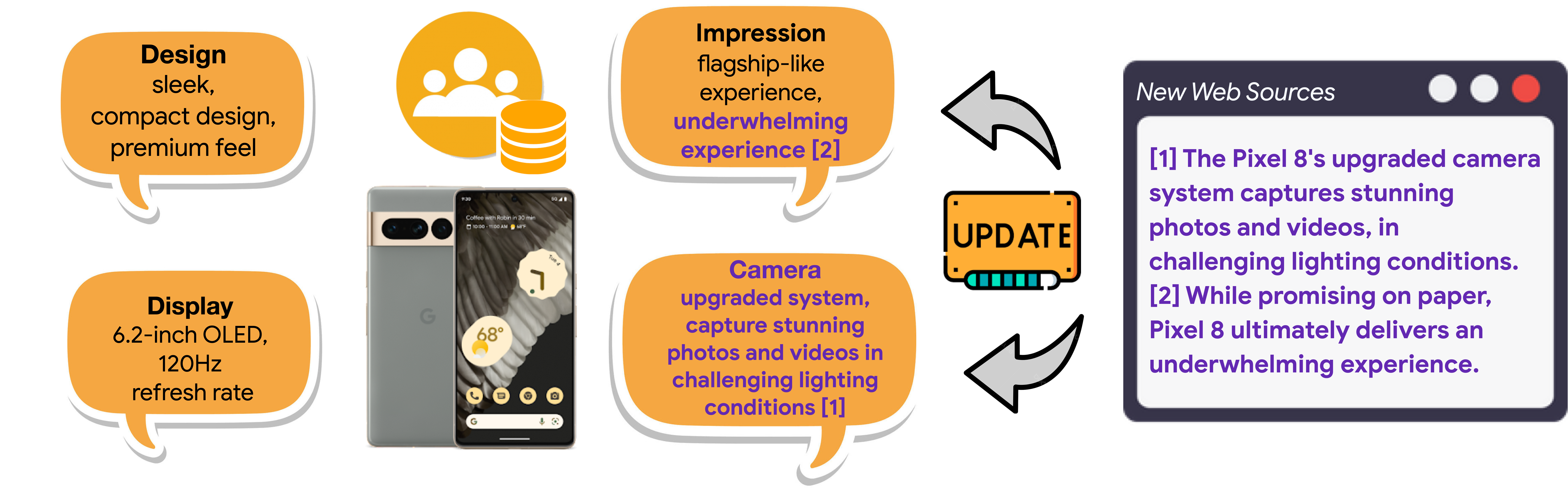}
    \caption{Overview of the Incremental Entity Summarization Task. Existing attribute (``Impression'') can be updated and new attribute (``Camera'') can be augmented.}
    \label{fig:task}
\end{figure*}

As the amount of information continues to grow, it is essential to be able to update these structured summaries automatically. %Incremental Entity Summarization (IES) solves this problem by letting models update entity summaries whenever new information becomes available~\citep{chowdhury2024incremental}. This ensures that search engines, news aggregators, and even knowledge bases always provide the most accurate and comprehensive information of key entities. Meanwhile, by eliminating manual updates, IES allows for efficient management of massive amounts of rapidly evolving information. As shown in Figure~\ref{fig:task}, more attributes and values can be updated to an existing entity summary table based on some new web sources, thereby progressively enhancing the comprehensiveness of entity representations.
Incremental Entity Summarization (IES) addresses this by enabling updates to entity summaries with new information~\citep{chowdhury2024incremental}, ensuring entities in search engines, news aggregators, and knowledge bases are always accurately represented. IES removes the need for manual updates, efficiently managing vast and swiftly changing data. It allows for the addition of attributes and values to entity summaries from new web sources, enhancing entity representation comprehensiveness as depicted in Figure~\ref{fig:task}.
Despite its critical importance, IES is underexplored. While some work~\citep{goasdoue2019incremental,yang2021incremental,chowdhury2024incremental} investigates updating entity summaries using abstractive or extractive techniques, these efforts often lack structured attribute-value organization or suffer from hallucination problems of LLMs. Crucially, there is no dataset specifically designed to test the ability of these models to maintain accurate, up-to-date entity knowledge. This is surprising given the vital role IES plays in organizing massive amounts of information.

%To build a dataset ideal for training IES systems with incremental generation capabilities, it should cover a wide range of entities with diverse attributes and values that vary in length, sentiment, and subjectivity. The dataset must include text sources exhibiting distinct writing styles to train the model's adaptability. Importantly, the dataset should simulate how information about entities evolves, forcing the model to update and revise its knowledge. Finally, a precise alignment between natural language web documents and their corresponding structured summaries is crucial, ensuring that each attribute value can be traced back to its exact textual source. 
To develop an effective dataset for IES systems, it needs a broad selection of entities with diverse and evolving attributes and values. This dataset should feature texts from varied writing styles to improve model adaptability and simulate the dynamic nature of entity information updates. Crucially, it requires accurate alignment between web documents and their structured summaries to trace attribute values to their sources. While diverse natural language web sources for various entities are readily available~\citep{ganesan2012opinion,asghar2016yelp}, creating well-maintained structured summaries from these sources remains both expensive and time-consuming. Even with the assistance of LLMs, achieving high accuracy and coverage in structured summary annotations is difficult and requires extensive human verification~\citep{gunel2023strum}. Further challenges arise in accurately representing how entity information evolves over time and maintaining a precise alignment between natural language and structured summaries~\citep{chowdhury2024incremental}. 

% Explain the source of inspiration and briefly describe the dataset generation procedures. Emphasize the overall main desiderata of the dataset.
In this paper, we strategically generate a fully synthetic dataset using LLMs. This approach leverages the empirical finding that LLMs excel at expanding short phrases into descriptive, contextual paragraphs, rather than abstractly summarizing all important components from longer text. %In detail, this paper employs a multi-step process to generate the dataset. First, popular search topics are used to guide LLM-based generation of diverse attributes and plausible entity names for each topic category. To ensure high-quality values for each attribute, the process requires descriptive values with varying lengths and sentiments. Additionally, to simulate real-world information updates, incremental summary tables are generated for each entity, with deliberate inclusion of repeated and conflicting information. Finally, to align with these summary tables, paragraphs are generated incorporating the attributes, values, and diverse writing styles with varied tones. 
The dataset generation follows a structured approach: initially, LLMs use popular search topics to produce varied attributes and plausible entity names. To ensure the quality of attribute values, which vary in length and sentiment, a meticulous process is applied. Additionally, the dataset mimics real-world updates by including incremental changes, conflicts, and repetitions in the entity summaries. Corresponding paragraphs that reflect these summaries are then generated, showcasing a range of writing styles and tones. Overall, this synthetic dataset is carefully crafted to be both high-quality and complex, enabling it to effectively simulate real-world scenarios.

% Contribution
Our contributions are as follows:
\begin{itemize}
    \item We present \datasetname, the first dataset built with high informativeness and diversity for rigorous evaluation of incremental entity summarization methods. We open-source \datasetname to accelerate research in this field, including metrics of evaluation.
    \item We propose simple but effective LLM-based solutions, \textbf{Update} and \textbf{Merge} for IES task. These methods provide valuable baselines for future advancements.
    \item We conduct insightful analyses to pinpoint the limitations of LLM-based entity summarization methods. State-of-the-art LLMs struggle to update summaries with an F1 score higher than 80.4\%, highlighting the inherent complexity of this task.
\end{itemize}
\section{Dataset Desiderata}
To build a dataset ideal for developing entity summarization systems with incremental generation capability, we outline the following key desiderata:

\noindent\textbf{Diversity of Entities.}
The dataset should encompass a broad spectrum of entities across domains. This could include businesses (restaurants, hotels), products, events, and more. Diverse entities ensure the model encounters a wide choices of attributes and associated values, expanding its knowledge base.

\noindent\textbf{Complexity of Attributes and Values.} 
Values associated with attributes should demonstrate variation in length, sentiment and subjectivity. Even within the same entity category, attribute values should reflect high diversity to challenge the models' nuanced understanding.
Likewise, attributes must range common (e.g. a restaurant's service) to niche and specific interests (e.g. a hiking trail's access to restrooms). 

\noindent\textbf{Varied Information Sources.}
The textual sources should exhibit a rich diversity of real-world styles and origins. Generate a mixture of editorial reviews (which often analyze with authority), user generated contents (informal and potentially biased, found in online forums or social media), and official product descriptions (which frequently use persuasive language focused on features and benefits). By exposing the model to these distinct writing styles and purposes, it will be compelled to adapt its understanding across different language patterns.

\noindent\textbf{Inclusion of Misleading Information}
The dataset should contain subtly misleading details that requires contextual understanding for identification. %This involves information that is factually accurate but becomes misleading when placed in the specific setting. 
The dataset can reference similar but distinct entities to create confusion. The goal is to challenge the model's ability to critically analyze information within the provided context rather than simply relying on basic fact-checking.

\noindent\textbf{Incremental Information Updates.}
The dataset should include examples where information about an entity evolves over time, simulating updates as new facets or perspectives are revealed. This forces the model to not only add new information but also potentially revise or re-prioritize existing facts. Introducing situations where initial information is incomplete or later contradicted by more supported sources. The model must learn to prioritize well-supported information over time, mirroring a common real-world scenario where our understanding of a subject develops.

\noindent\textbf{Rigorous Alignment between Structured Summaries and Natural Language Paragraphs.}
Ensure a precise and traceable connection exists between a source paragraph and its corresponding structured summary (i.e. an attribute-value table). Focus on maintaining clear attributions, and ensure the origin of each value is precisely derived from the source paragraph. Avoid introducing information into the structured summary that isn't explicitly supported by the text for a rigorous alignment. % This rigorous alignment is essential for the model to learn to pinpoint the exact evidence for specific attribute values, reducing the potential for misinterpretation.

\section{Dataset Generation Methodology} 
\begin{figure*}[t]
    \centering
    \vspace{-30pt}
    \includegraphics[width=\linewidth]{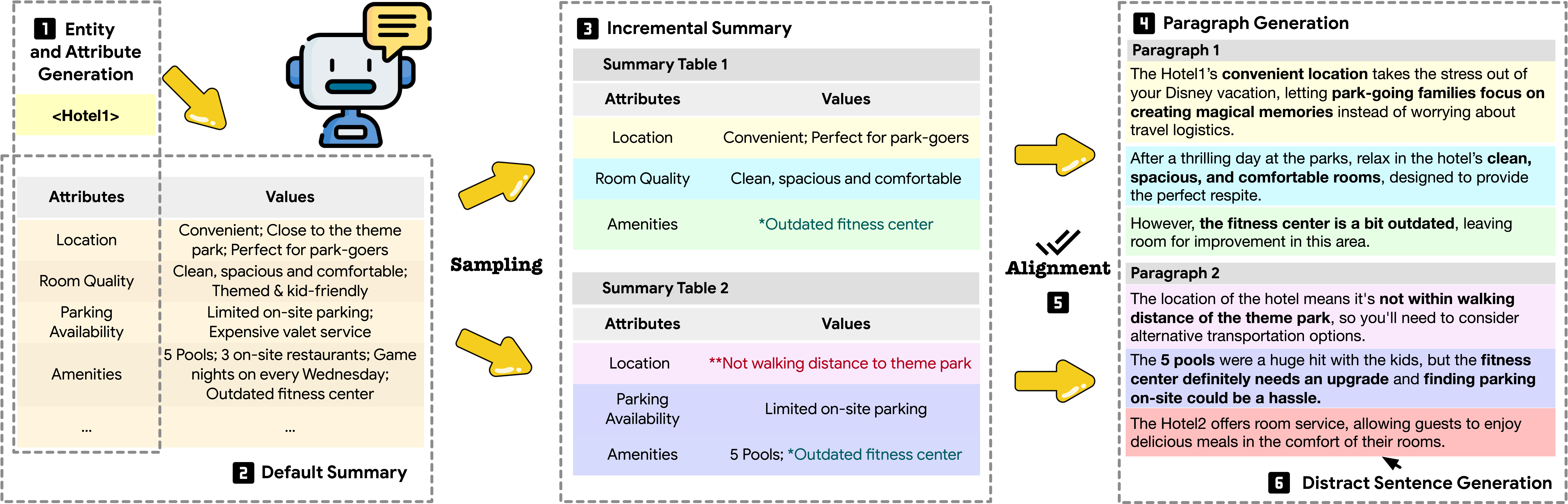}
    \caption{Dataset Generation Methodology Overview: (1) Generate entity names (masked for ethics consideration) and attributes. (2) Create default summary table with diverse values. (3) Sample attributes/values for incremental summaries (* repeated sampling, ** conflicting values). (4) Generate paragraphs with varying tones based on attributes/values. (5) Verify summary table/paragraph alignment. (6) Add distractor sentence. Note that attribute values and sentences in the same color should be aligned and bold texts in paragraphs are the evidences for corresponding attribute values.}
    \label{fig:flow}
\end{figure*}

We create a synthetic dataset with generated attributes, entity names, and incrementally evolving summary tables (see Figure~\ref{fig:flow}). Accompanying paragraphs mirror the tables, including distracting sentences. For LLM prompting instructions, see Appendix~\ref{app:dataset-generation-prompts}.

\subsection{Attribute and Entity Name Generation}
We begin by selecting 20 popular categories (e.g. \texttt{Hotels \& Accomodations}) (see Appendix \ref{app:dataset-stats} for all category information). For each, we prompt an LLM to generate attributes (e.g. \texttt{Room Quality}) and entity names (e.g. \texttt{The Whispering Canyon Hotel}). To ensure attribute diversity, we retrieve up to 50 common (e.g. \texttt{Room quality}, \texttt{Service}) and 50 less-common attributes (e.g. \texttt{Honeymoon packages}) typically used to describe entities within that category. For entity names, we generate up to 40 plausible but fictitious names, randomly selecting 10. Each entity is then assigned 30 attributes, with an equal split between common and uncommon descriptors. This process results in a dataset containing 200 entities, which we consider \textit{suitable} for evaluation. The use of random elements in the generation process helps reduce the impact of LLM bias on the dataset. While we use specific entity names to promote contextual diversity in the models' output until we generate paragraphs (Sec~\ref{sec:paragraph}), in the final dataset, these names are replaced with generic ones (e.g., from \texttt{The Whispering Canyon Hotel} to \texttt{Hotel1}) to avoid any unintended claims related to real-world entities. 

\subsection{Summary Table Generation}
\label{sec:summary-table-gen}
\paragraph{Default summary table generation.} Summary tables provide a structured representation of attributes associated with an entity in a given category. Each row details an attribute and its corresponding value. The goal in this stage is to generate values that meet three criteria: 1) Informative and meaningful, covering both subjective and objective aspects, 2) Diverse in length (one to 10 words), and 3) Varied in sentiment (positive, negative, and neutral). We generate at least three descriptive values per sentiment, resulting in three distinct summary tables for each entity. For instance, when the prompt specifies a positive sentiment, the model is directed to generate favorable descriptors such as ``\texttt{Spacious and comfortable}'' and ``\texttt{Clean}'' for a designated attribute like ``\texttt{Room Quality}''. The final summary tables for each entity combine up to 10 attribute and value pairs, including varied sentiments derived from 3 separate summaries for each entity.

\paragraph{Incremental summary table generation.} To assess the LLM's incremental update capabilities, we generate multiple summary tables per entity. The initial summary is the basis from which we sample attributes and values for incremental versions. These incremental summaries simulate real-world scenarios where information evolves. We ensure two criteria are met: 1) Repetition of attributes and values across summaries, and 2) The presence of conflicting attribute information. Conflicting values can be generated by prompting an LLM to produce values that directly oppose the meanings of originally sampled values. This incremental summary generation iteratively creates $K$ summaries. Each iteration combines half the attributes from a previous summary with half from the unused attribute pool, resulting in $K$ summary tables per entity with diverse and potentially contradictory content.

\subsection{Paragraphs}
\label{sec:paragraph}
\paragraph{Paragraph generation.}
Building upon the $K$ incrementally generated summary tables (Section~\ref{sec:summary-table-gen}), we craft aligned paragraphs for each. The fundamental goal is to seamlessly incorporate all attributes and values from a given table into the text. Additionally, we prioritize meaningful and diverse writing styles, avoiding overly simplistic language. To achieve this, we define 8 writing categories (user reviews by teenagers, user reviews by parents, user reviews by senior, user reviews, official product descriptions, editorial insights, posts on social media, discussions on online forums) and 6 tones (optimistic, neutral, pessimistic, sarcastic, humorous, analytic). Each paragraph is randomly assigned a category and a tone, which guide its generation. We also integrate citation numbers (e.g. 0, 1, 2, ...) that directly link each sentence to the attribute-value pair it reflects in the summary table. This process results in $K$ paragraphs per entity (in our case, $K$=7), showcasing a variety of styles, tones, and embedded citations for easy reference.

\paragraph{Paragraph-Summary table alignment verification.} While the sentences in paragraphs are created based on summary tables, the generated paragraphs do not guarantee that all values are reflected in sentences. To make sure that sentences include the attribute-value pairs in the given summary table, we break paragraphs down into sentences and LLM verifies if the attribute-value pair (e.g. (\texttt{Room quality}, \texttt{Clean})) is accurately represented in each sentence. If the value is correctly included and its meaning is not misrepresented (e.g. \texttt{With its impeccable clean rooms...}), no change is needed. If the value is missing or misrepresented (e.g. \texttt{While its cleanliness of the rooms are debatable...}), the sentence should be adjusted to accurately incorporate the value.

After the automated critique and revision step, we conduct human evaluation on 40 randomly sampled paragraphs and their corresponding summary tables. Three human annotators checked for misaligned attribute-value pairs in the paragraphs based on the summary tables. Our dataset achieved 96\% of accuracy with 70\% of agreement rate.\footnote{It means 70\% of examples demonstrate full agreement among three annotators. Furthermore, 100\% examples achieve at least two-annotator agreement.}

\subsection{Distracting Sentences}
After ensuring paragraph-summary table alignment, where all sentences contain attribute-value pairs, we introduce distractor sentences to test the LLM's focus. Since LLMs perform well in finding relevant contexts, we need to challenge their ability to identify and ignore incorrect entity associations. We do this in two ways: first, by generating sentences about irrelevant entities, explicitly including their generic names (e.g. \texttt{HOTEL2 boasts a vibrant atmosphere, perfect for...}), and second, by creating metaphorical sentences that describe a human using properties of the given entity's category (always including the word "HUMAN") (e.g. \texttt{HUMAN's empathy is a sprawling garden, teeming with vibrant blooms of compassion...}). These distractors allow us to analyze two crucial aspects of LLM performance: entity focus (avoiding irrelevant information) and adjective sensitivity (understanding adjectives even in unrelated contexts).

\subsection{Dataset Statistics}
\label{sec:data-stat}
We present our dataset statistics for the entity level and paragraph level in Appendix \ref{app:dataset-stats}. Overall, the dataset contains 200 entities (20 for each of the 10 categories) and each entity contains an average of 22 attributes and 42 values across all paragraphs, which we believe, achieves \textit{sufficient complexity} for evaluation.
Entities within the same category display a significant amount of diversity. They have approximately 14 distinct attributes (64\%) and 41 distinct values (97\%) on average. This demonstrates a high degree of variation in their attributes and values.
In paragraph statistics (in Appendix \ref{app:dataset-stats}), we find that number of ``same'', ``conflict'', and ``new'' attribute values in each paragraph are around 3.7, 3.5, and 2.3, respectively, meaning that same, conflict, and new attribute-value pairs are reasonably distributed across paragraphs. Average number of sentences in paragraphs is 12, with roughly 4 sentences acting as distractors. This indicates that our paragraphs offer sufficient length and incorporate a reasonable amount of distractor sentences. 

We show 5 dataset examples in the Appendix from Figure~\ref{fig:example1} to \ref{fig:example5} in categories of ``Computer \& Video Games'', ``Vitamins \& Supplements'', ``Restaurants \& Bars'', ``Books \& Literature'', and ``Education'', respectively.
\section{Experiments}

\subsection{Baseline Methods}
Our dataset evaluation utilizes two prompt-based approaches, \update and \merge, designed to assess the LLM's ability to handle new information and conflicts. 
% \paragraph{Generate.} The model is given with a paragraphs associated with the specific entity. Its task is to generate a new comprehensive summary table by understanding all paragraphs, and by identifying all new, conflicting, or matching information from the long paragraphs.

\paragraph{Update.} LLMs struggle to create comprehensive, high-quality summary tables from large amounts of text due to limited recall~\citep{gunel2023strum}. We address information overload and reduce the LLM's processing burden by feeding it one paragraph at a time. The first iteration involves generating a summary table from a single paragraph. Afterwards, the LLM receives a new paragraph (potentially containing overlapping, new, or conflicting information) and the previously generated summary table. Its goal is to produce an updated summary table, accurately incorporating relevant details from the new paragraph. Prompts for this method can be found in Appendix~\ref{app:eval-prompts}.

\paragraph{Merge.} This approach breaks down the \update process into two steps, designed to enhance the LLM's understanding. The first iteration remains the same as the \update, with the model generating a summary table from a new paragraph. In later iterations, the model first creates a summary table solely from the new paragraph and then merges it with the existing table. This promotes a clear understanding of the two-step process of retrieving information and updating the summary, potentially reducing the LLM's cognitive load. Prompts for this method can be found in Appendix~\ref{app:eval-prompts}.

\subsection{Evaluation Metrics} 
\label{sec:eval-metrics}
We evaluate the performance of the aforementioned approaches to the incremental entity summarization task using precision, recall, and F1. An extraction comprises three components -- the attribute, its corresponding value, and the supporting evidence. A successful extraction is one that is also found in the set of goldens corresponding to the input paragraph.

We determine \textit{true positives} via two methods.
\textbf{Exact matching} checks for a direct match between the predicted value or evidence and the golden set.\footnote{But in tasks like structured entity summarization, it is rare that extractive techniques exactly match the gold truth sets, as such we need methods that can handle the nuance of language.}
\textbf{LLM-based evidence finding} leverages an LLM to detect if the predicted attribute and value find support within the larger golden set (see Appendix~\ref{app:eval-prompts} for prompt). 
Initially, we explored semantic similarity scores like BLEURT~\citep{bleurt}. However, their limitations in handling paraphrases led us to adopt an LLM for this task, as it excels at identifying evidence even when phrased differently.
If a predicted extraction fails to match exactly or through the LLM-based evidence prompt, it's marked as a \textit{false positive}. \textit{False negatives} are tracked by noting goldens unmatched to any prediction. While exact matches are simple, LLM-based matches are trickier. The LLM outputs the matched golden row (attribute, value, evidence), but it may not precisely align with the table due to the LLM's generative nature. To address this, we evaluate the cosine similarities between a sentence encoding (we use Universal Sentence Encoder~\citep{cer2018universal}) of the response's evidence to the sentence encodings of all the evidences in the golden set to find the highest likelihood golden.

To check its effectiveness in identifying evidence linking predicted and gold-standard attribute values, we manually checked up to 3 paragraphs under the 3 categories, which include more than 210 attribute-value pairs to evaluate. We count incorrectly classified pairs in true positive, false positive, and false negative sets. The \pro model achieves 90.4\% accuracy in evidence detection with a standard deviation of 1\% across categories, proving its suitability as an evidence detector between predicted and gold values.

Redundancy and hallucination are crucial metrics requiring evaluation. Redundancy, where models repeatedly extract the same correct value, can artificially inflate F1 scores and hinder fair performance comparisons. Moreover, LLMs are prone to hallucinations, where they generate incorrect values from extracted evidences. Though these hallucinations negatively impact precision and F1 scores, we still want to explicitly measure its severity. For a thorough analysis, we employed two human experts to manually assess these issues within the predicted summary tables; their findings are discussed in the Section~\ref{sec:results}.

\begin{table}[t]
\scriptsize
\centering
\vspace{-30pt}
\begin{tabular}{l|l|l|rrrrrrr|r}
\toprule
 &  & \multicolumn{7}{c}{Turns} \\

 & Model & Metric & 1 & 2 & 3 & 4 & 5 & 6 & 7 & Avg.\\
\midrule
% \hline
% \multirow{12}{*}{UD} & Gemini-Ultra & Precision & 77.5 & 75.7 & 75.9 & 76.9 & 79.1 & 79.9 & 81.0 & 78.0 \\
% &  & Recall & 83.4 & 80.2 & 79.1 & 73.2 & 66.8 & 60.9 & 56.3 & 71.4 \\ 
% & & F1 & 79.2 & 76.8 & 76.4 & 73.2 & 70.3 & 67.5 & 64.9 & 72.6 \\
% \cmidrule{2-11}

 \multirow{9}{*}{UD} & Gemini-Pro & Precision & 80.0 & 81.9 & 82.6 & 82.5 & 83.8 & 84.1 & 84.3 & 82.8 \\
&  & Recall & 82.5 & 76.2 & 73.2 & 70.4 & 69.7 & 68.4 & 67.3 & 72.5 \\ 
& & F1 & \textbf{80.7} & \textbf{78.4} & \textbf{77.2} & 75.3 & 75.5 & 74.8 & 74.2 & \textbf{76.6} \\
 \cmidrule{2-11}

& GPT3.5 & Precision & 78.7 & 78.2 & 79.3 & 79.4 & 79.8 & 79.7 & 80.0 & 79.3 \\
&  & Recall & 81.6 & 78.1 & 75.7 & 74.8 & 74.8 & 74.9 & 75.1 & 76.4 \\ 
&  & F1 & 79.5 & 77.6 & 77.0 & \textbf{76.7} & \textbf{76.8} & \textbf{76.9} & \textbf{77.1} & 77.4 \\

 \cmidrule{2-11}

& Gemini-Nano & Precision & 58.7 & 52.6 & 49.0 & 47.0 & 46.1 & 46.0 & 45.9 & 49.3\\
&  & Recall & 65.4 & 43.5 & 31.0 & 25.0 & 21.1 & 18.6 & 16.2 & 31.5 \\ 
&  & F1 & 60.7 & 46.4 & 37.2 & 31.9 & 28.3 & 26.0 & 23.5 & 36.3 \\

\bottomrule
\toprule
%  Gemini-Ultra & Precision & 76.4 & 77.0 & 77.6 & 78.4 & 79.8 & 80.2 & 81.6 & 78.7 \\
%  & & Recall & 82.7 & 80.5 & 81.9 & 78.0 & 71.6 & 63.7 & 60.0 & 74.1 \\ 
%  & & F1 & 78.8 & 76.7 & 78.6 & 77.0 & 74.2 & 69.7 & 68.1 & 74.7 \\
% \cmidrule{2-11}

\multirow{9}{*}{MG} & Gemini-Pro & Precision & 79.4 & 79.7 & 80.4 & 80.6 & 80.8 & 81.8 & 83.1 & 80.8 \\
&  & Recall & 82.1 & 84.0 & 83.2 & 82.0 & 81.1 & 78.7 & 74.8 & 80.8 \\ 
& & F1 & \textbf{80.1} & 81.4 & \textbf{81.4} & \textbf{80.9} & \textbf{80.5} & \textbf{79.9} & 78.3 & \textbf{80.4} \\
 \cmidrule{2-11}

& GPT3.5 & Precision & 75.7 & 76.4 & 76.1 & 77.4 & 76.4 & 76.2 & 77.7 & 76.6 \\
&  & Recall & 83.3 & 88.3 & 87.8 & 85.3 & 84.2 & 82.9 & 82.5 & 84.9 \\ 
&  & F1 & 78.8 & \textbf{81.6} & 81.3 & 80.8 & 79.8 & 79.1 & \textbf{79.8} & 80.2 \\
 \cmidrule{2-11}

& Gemini-Nano & Precision & 60.0 & 51.0 & 53.7 & 54.0 & 56.7 & 57.9 & 57.1 & 55.8 \\
&  & Recall & 66.5 & 47.4 & 37.0 & 32.0 & 29.6 & 25.5 & 22.0 & 37.1 \\ 
&  & F1 & 62.1 & 47.9 & 42.1 & 38.4 & 37.5 & 33.9 & 30.5 & 41.8 \\
  
% &  & Redundancy & 0.738 & 0.807 & 0.771 & 0.754 & 0.783 & 0.768 & 0.768 \\
% &  & Hallucination & 0.738 & 0.807 & 0.771 & 0.754 & 0.783 & 0.768 & 0.768 \\
% \cmidrule{3-10}
%  & & Avg. & 0.738 & 0.807 & 0.771 & 0.754 & 0.783 & 0.768 & 0.768 \\

\bottomrule
\end{tabular}

\caption{Performance with \pro, \chatgpt, and \nano models across different turns. UD denote \update and MG denote \merge. Best F1 scores are in \textbf{boldface}.}
\label{tab:eval-results}
\end{table}

\subsection{Experimental Setup}
We experiment with \pro~\citep{gemini}, \chatgpt~\citep{openai}, and \nano~\citep{gemini} models. 
The temperatures for all models are set to 0.7. With each entity having 7 paragraphs, we aggregate summary tables iteratively, reporting average precision, recall, and F1 scores across all entities. 

\subsection{Results and Discussion}\label{sec:results}

\paragraph{Overall performance, Table \ref{tab:eval-results}.} 
% first discuss f1 scores in later iterations and mention precision and recall tradeoff
Table \ref{tab:eval-results} shows an overall performance of \pro, \chatgpt, and \nano on our dataset. At first glance, all models have a large room for improvement, highlighting our dataset's complexity. In particular, \nano model performs significantly worse than \pro and \chatgpt in both \update and \merge methods, with an average F1 score gap of 40.3 for \update and 38.6 for \merge.
The reason of this substantial performance gap between \pro/\chatgpt and \nano models can be attributed to the differences across iterations. While the \nano model performs reasonably well in the first iteration, its recall scores begin to rapidly decline from the second iteration, resulting in a decrease of more than 20 points. By the final iteration, the \nano model produces significantly lower scores, reaching 23.5 for \update and 30.5 for \merge. This suggests that as the context becomes more complex, smaller LLMs struggle to identify and integrate new or relevant information effectively. Moreover, when the input context in the prompt becomes overly lengthy, \nano model struggles to understand instructions correctly, leading to up to 13\% of invalid answers, such as repeating the input prompt in the response. In contrast, \pro and \chatgpt generate substantially fewer invalid answers (around 0\%) even with complex inputs.

Interestingly, while \pro and \chatgpt show comparable performance, \pro tends to produce higher precision scores, suggesting that it prioritizes confident and accurate answers. On the other hand, \chatgpt achieves better recall scores, indicating that it explores a broader range of answer choices. This becomes more evident in later iterations.
While \chatgpt model produces relatively stable performance in both precision and recall scores across all iterations, Gemini models exhibit a trade-off between precision and recall scores. This shows that Gemini models prioritize to keep the reliable results with the complex context. 
None of these advanced LLMs achieved higher F1 than 80.4\%, supporting the empirical finding that LLMs excel at expanding short phrases into descriptive, contextual paragraphs, rather than abstractly summarizing all important components from longer text.

\paragraph{Difference across methods, Table \ref{tab:eval-results}.} We observe that models perform better with \merge method than \update approach. This confirms our hypothesis that breaking down \update method into two steps gives a better understanding of our task to LLMs. The \merge method is particularly beneficial for maintaining recall scores. This is likely because it first extracts attributes and values from the given new paragraph, which are then presented to the model for merging with the existing knowledge. By making the information we want to add explicit in the prompt, the model can more easily make use of the given knowledge.

\begin{figure}[t]
\small
\vspace{-30pt}
\includegraphics[clip,width=1\linewidth]{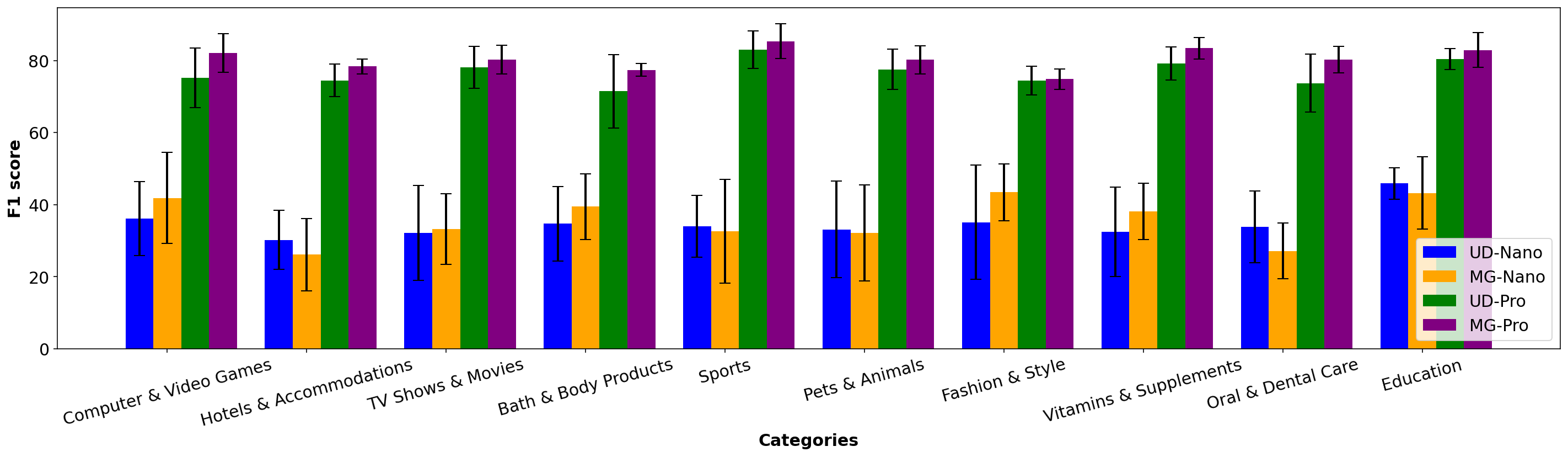}
% \subfloat[10 categories from Computer \& Video Games to Education]{%
%   \includegraphics[clip,width=1\linewidth]{Figures/category_performance_0_10.png}%
% }

% \subfloat[Remaining 10 categories from Fruits \& Vegetables  to Books \& Literature]{%
%   \includegraphics[clip,width=1\linewidth]{Figures/category_performance_10_20.png}%
% }

\caption{F1 scores across 10 categories (see Appendix \ref{app:performance-chart} for the rest.).}
\label{fig:performance-cate}
\end{figure}

\paragraph{Difference across categories and tones, Figure \ref{fig:performance-cate}, \ref{fig:performance-tone}.}
Figure~\ref{fig:performance-cate} presents the F1 scores achieved by the model across different categories, along with their standard deviations. As the figure shows, the model exhibits consistent performance across all categories. There are no significant outliers, implying that the performance of models on our dataset is not biased towards certain categories. Figure~\ref{fig:performance-tone} in the Appendix shows the performance across paragraph tones and we observe the similar trends to the performance across categories. We also note that standard deviations of \nano models are considerably larger than those of \pro models in most cases, reconfirming the challenging nature of our dataset.

\begin{table}[t]
\scriptsize
\centering
\begin{tabular}{l|l|rrrrrrr|r}
\toprule
& & \multicolumn{7}{c}{Turns} & \\

 & Metric & 0 & 1 & 2 & 3 & 4 & 5 & 6 & Avg.\\
\midrule
w/ distractor &  Precision & 80.0 & 81.9 & 82.6 & 82.5 & 83.8 & 84.1 & 84.3 & 82.8 \\
& Recall & 82.5 & 76.2 & 73.2 & 70.4 & 69.7 & 68.4 & 67.3 & 72.5 \\ 
& F1 & 80.7 & 78.4 & 77.2 & 75.3 & 75.5 & 74.8 & 74.2 & 76.6 \\
\midrule
w/o distractor & Precision & 97.2 & 96.5 & 96.6 & 96.6 & 97.0 & 96.7 & 96.8 & 96.8 \\
& Recall & 84.3 & 77.8 & 74.3 & 72.8 & 71.5 & 70.5 & 70.1 & 74.5 \\ 
& F1 & 89.9 & 85.8 & 83.7 & 82.7 & 82.0 & 81.3 & 81.0 & 83.8 \\
\bottomrule
\end{tabular}

\caption{Precision, Recall, and F1 score after removing distractor sentences. }
\label{tab:performance-dist}
\end{table}

\begin{figure*}[t]
\small
    \centering
    \vspace{-30pt}
    \includegraphics[width=1\linewidth]{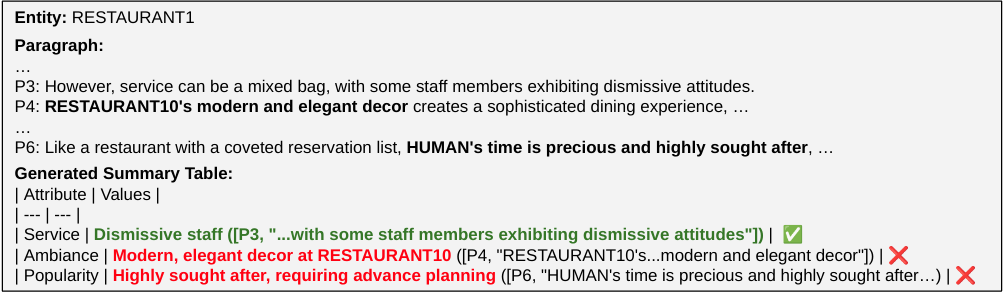}
    \caption{An example of an LLM distracted by irrelevant information.}
    \label{fig:dist-effect}
\end{figure*}

\paragraph{Effect of distractor sentences, Table \ref{tab:performance-dist}, Figure \ref{fig:dist-effect}.} Table \ref{tab:performance-dist} shows the performance of \pro model with \update method after removing distractor sentences in paragraphs. We find that precision scores achieve up to 97 point when the distractor sentences are removed. This proves that our distractor sentences are effectively confusing LLMs and LLMs struggle in strictly focusing on the context relevant to the specific entity. Moreover, it further indicates that our evaluation method based on LLMs works reasonably well in detecting evidence between generated attribute and value pairs and gold attribute, value, and sentence pairs. Figure \ref{fig:dist-effect} shows an example of incorrect output from LLMs with distractor sentences. We find that LLMs can easily be misled by information that include several adjective words and also struggle in distinguishing context crucial to the specific entity.

\paragraph{Human evaluation for checking value redundancy.}
In addition to F1 scores, we perform two human evaluations to assess how well the model consolidates similar attribute-value pairs (redundancy checking) and to check how well the extracted evidence supports the values. For redundancy checking, two annotators are presented with 30 randomly selected attributes with more than two distinct values generated by \pro. They indicate `yes' if the values for each attribute are redundant (e.g., Attribute: \texttt{Location}, Values: [\texttt{Walking distance from Downtown}, \texttt{easy access to Downtown}]), and 'no' otherwise. 
This evaluation is crucial because an excessive number of similar values for the same attribute can inflate true positives, resulting in artificially high precision and recall scores. We find that, on average, 45\% of values are deemed redundant with a 73\% agreement rate, indicating that the LLM struggles with identifying and merging synonyms into a single value.

\paragraph{Human evaluation for hallucination between value and evidence.} 
Similarly to redundancy checking, two annotators are tasked with assessing the alignment between extracted evidence and attribute values. They are provided with 30 randomly selected attributes, along with their corresponding values and evidences. The annotators mark `yes' if the attribute and values are supported by evidence, and `no' otherwise. 
This allows us to assess the faithfulness of LLMs in extracting evidences to support attribute and values. We find that an average of 25\% of the samples are marked as `no', meaning that evidence does not support the values, with 90\% of agreement ratio. An example where the value is not supported by evidence is the attribute-value pair ``Guest Privileges'' with the value ``ability to earn points that can be redeemed for free nights,'' and the evidence provided by the model is ``Your loyalty will be rewarded,'' where the evidence does not explicitly mention earning points or free nights. This suggests that LLMs often generate attributes and additional details that are not directly supported by the source information.

\section{Related Work}
Our review delves into incremental entity summarization (IES), analyzing approaches, methodologies, and datasets, with a particular emphasis on knowledge update techniques and conflict resolution.

\subsection{Techniques for Incremental Entity Summarization}
Current ES research has largely focused on summarizing entities from RDF data by selecting key triples~\citep{wei2019esa,liu2020entity,liu2021entity}, aiming for compact summaries. 
Our approach, in contrast, seeks to harness unstructured web text for more comprehensive summaries. While Formal Concept Analysis shows promise in structured knowledge bases~\citep{yang2021incremental}, it struggles with the complexity of web information. 
Existing datasets~\citep{liu2020esbm,gunaratna2015faces,gunaratna2016gleaning} fall short in testing LLMs' capabilities for web-driven, incremental summary generation. 
The ENTSUM dataset~\citep{maddela2022entsum} aids in controllable summarization but is limited in assessing structured or incremental summary creation.
Our work broadens the definition of IES, investigating the construction of comprehensive and precise structural summaries using advanced generation models such as LLMs, and introducing a dataset tailored for this innovative field.

\subsection{Addressing Knowledge Updates and Conflicts}
%A core challenge in incremental entity summarization is ensuring LLMs can identify knowledge updates and resolve conflicts. While tools like CoverSumm~\citep{chowdhury2024incremental} streamline summary updates, and datasets like KNOWLEDGE CONFLICT~\citep{wang2023resolving} assess LLM conflict handling by creating examples simply by either substituting named entities within the same domain or shuffling main entities between contexts in the same domain, introducing knowledge conflicts. However, they lack the complexity and adaptability required for incremental entity summarization. FreshQA~\citep{vu2023freshllms}, a QA benchmark, though valuable for evaluating factuality, does not directly focus on evolving entity summaries. Our dataset addresses this gap, uniquely requiring LLMs to detect conflicts within entity summaries and dynamically reprioritize claims based on the strength of their supporting evidence, mirroring the demands of incremental entity summarization.
The main challenge in IES is enabling LLMs to handle knowledge updates and resolve information conflicts. Solutions like CoverSumm~\citep{chowdhury2024incremental} and the KNOWLEDGE CONFLICT dataset~\citep{wang2023resolving} offer ways to update summaries and test conflict resolution, albeit without the necessary complexity for IES. Similarly, the FreshQA benchmark~\citep{vu2023freshllms} tests LLM factuality but doesn't cater specifically to evolving summaries. Our dataset fills this gap, demanding LLMs to identify and adjust to conflicts in entity summaries, with a focus on evidence-based claim reprioritization, aligning closely with the unique requirements of IES.

%\subsection{LLM-based Synthetic Datasets}

\section{Conclusion and Future Work}
In this paper, we introduce \datasetname, a novel benchmark, specifically created to assess the ability of LLMs to generate incremental summaries of entities. \datasetname's synthetic nature ensures data quality and diversity while minimizing the need for extensive human annotations. We also share our thoughts about paper's limitation in the Appendix~\ref{app:limitation}.
While our initial baselines demonstrate the dataset's challenges, future work offers exciting avenues. We will focus on preventing knowledge loss during LLM updates, refining attribute and value recognition to minimize hallucinations, and extending the task to multi-entity comparison summaries. 
%Additionally, updating real-world knowledge graphs or knowledge bases would introduce the challenge of managing relationships alongside individual attribute updates.
Overall, this paper aims to spark impactful research into the crucial task of maintaining up-to-date and comprehensive knowledge.

\section{Ethics Statement}
Our dataset is primarily meant to serve as a diagnostic tool to evaluate LLMs' ability of resolving knowledge conflicts incrementally and generating faithful responses. In addition, the LLMs we used for creating the dataset are trained on a large-scale web corpus and may also bring some bias when generating sentences.

\bibliography{colm2024_conference}

\begin{thebibliography}{21}
\providecommand{\natexlab}[1]{#1}
\providecommand{\url}[1]{\texttt{#1}}
\expandafter\ifx\csname urlstyle\endcsname\relax
  \providecommand{\doi}[1]{doi: #1}\else
  \providecommand{\doi}{doi: \begingroup \urlstyle{rm}\Url}\fi

\bibitem[Allam \& Haggag(2012)Allam and Haggag]{allam2012question}
Ali Mohamed~Nabil Allam and Mohamed~Hassan Haggag.
\newblock The question answering systems: A survey.
\newblock \emph{International Journal of Research and Reviews in Information Sciences (IJRRIS)}, 2\penalty0 (3), 2012.

\bibitem[Asghar(2016)]{asghar2016yelp}
Nabiha Asghar.
\newblock Yelp dataset challenge: Review rating prediction.
\newblock \emph{arXiv preprint arXiv:1605.05362}, 2016.

\bibitem[Cer et~al.(2018)Cer, Yang, yi~Kong, Hua, Limtiaco, John, Constant, Guajardo-Cespedes, Yuan, Tar, Sung, Strope, and Kurzweil]{cer2018universal}
Daniel Cer, Yinfei Yang, Sheng yi~Kong, Nan Hua, Nicole Limtiaco, Rhomni~St. John, Noah Constant, Mario Guajardo-Cespedes, Steve Yuan, Chris Tar, Yun-Hsuan Sung, Brian Strope, and Ray Kurzweil.
\newblock Universal sentence encoder, 2018.

\bibitem[Chowdhury et~al.(2024)Chowdhury, Monath, Dubey, Zaheer, McCallum, Ahmed, and Chaturvedi]{chowdhury2024incremental}
Somnath Basu~Roy Chowdhury, Nicholas Monath, Avinava Dubey, Manzil Zaheer, Andrew McCallum, Amr Ahmed, and Snigdha Chaturvedi.
\newblock Incremental extractive opinion summarization using cover trees.
\newblock \emph{arXiv preprint arXiv:2401.08047}, 2024.

\bibitem[Ganesan \& Zhai(2012)Ganesan and Zhai]{ganesan2012opinion}
Kavita Ganesan and ChengXiang Zhai.
\newblock Opinion-based entity ranking.
\newblock \emph{Information retrieval}, 15\penalty0 (2):\penalty0 116--150, 2012.

\bibitem[Goasdou{\'e} et~al.(2019)Goasdou{\'e}, Guzewicz, and Manolescu]{goasdoue2019incremental}
Fran{\c{c}}ois Goasdou{\'e}, Pawel Guzewicz, and Ioana Manolescu.
\newblock Incremental structural summarization of rdf graphs.
\newblock In \emph{EDBT 2019-22nd International Conference on Extending Database Technology}, 2019.

\bibitem[Gunaratna et~al.(2015)Gunaratna, Thirunarayan, and Sheth]{gunaratna2015faces}
Kalpa Gunaratna, Krishnaparasad Thirunarayan, and Amit Sheth.
\newblock Faces: diversity-aware entity summarization using incremental hierarchical conceptual clustering.
\newblock In \emph{Proceedings of the AAAI Conference on Artificial Intelligence}, volume~29, 2015.

\bibitem[Gunaratna et~al.(2016)Gunaratna, Thirunarayan, Sheth, and Cheng]{gunaratna2016gleaning}
Kalpa Gunaratna, Krishnaprasad Thirunarayan, Amit Sheth, and Gong Cheng.
\newblock Gleaning types for literals in rdf triples with application to entity summarization.
\newblock In \emph{The Semantic Web. Latest Advances and New Domains: 13th International Conference, ESWC 2016, Heraklion, Crete, Greece, May 29--June 2, 2016, Proceedings 13}, pp.\  85--100. Springer, 2016.

\bibitem[Gunel et~al.(2023)Gunel, Tata, and Najork]{gunel2023strum}
Beliz Gunel, Sandeep Tata, and Marc Najork.
\newblock Strum: Extractive aspect-based contrastive summarization.
\newblock In \emph{Companion Proceedings of the ACM Web Conference 2023}, pp.\  28--31, 2023.

\bibitem[Kowalski(2007)]{kowalski2007information}
Gerald~J Kowalski.
\newblock \emph{Information retrieval systems: theory and implementation}, volume~1.
\newblock springer, 2007.

\bibitem[Liu et~al.(2020{\natexlab{a}})Liu, Chen, Cheng, Kharlamov, Li, and Qu]{liu2020entity}
Qingxia Liu, Yue Chen, Gong Cheng, Evgeny Kharlamov, Junyou Li, and Yuzhong Qu.
\newblock Entity summarization with user feedback.
\newblock In \emph{The Semantic Web: 17th International Conference, ESWC 2020, Heraklion, Crete, Greece, May 31--June 4, 2020, Proceedings 17}, pp.\  376--392. Springer, 2020{\natexlab{a}}.

\bibitem[Liu et~al.(2020{\natexlab{b}})Liu, Cheng, Gunaratna, and Qu]{liu2020esbm}
Qingxia Liu, Gong Cheng, Kalpa Gunaratna, and Yuzhong Qu.
\newblock Esbm: an entity summarization benchmark.
\newblock In \emph{The Semantic Web: 17th International Conference, ESWC 2020, Heraklion, Crete, Greece, May 31--June 4, 2020, Proceedings 17}, pp.\  548--564. Springer, 2020{\natexlab{b}}.

\bibitem[Liu et~al.(2021)Liu, Cheng, Gunaratna, and Qu]{liu2021entity}
Qingxia Liu, Gong Cheng, Kalpa Gunaratna, and Yuzhong Qu.
\newblock Entity summarization: State of the art and future challenges.
\newblock \emph{Journal of Web Semantics}, 69:\penalty0 100647, 2021.

\bibitem[Maddela et~al.(2022)Maddela, Kulkarni, and Preotiuc-Pietro]{maddela2022entsum}
Mounica Maddela, Mayank Kulkarni, and Daniel Preotiuc-Pietro.
\newblock Entsum: A data set for entity-centric summarization.
\newblock \emph{arXiv preprint arXiv:2204.02213}, 2022.

\bibitem[Ouyang et~al.(2022)Ouyang, Wu, Jiang, Almeida, Wainwright, Mishkin, Zhang, Agarwal, Slama, Ray, Schulman, Hilton, Kelton, Miller, Simens, Askell, Welinder, Christiano, Leike, and Lowe]{openai}
Long Ouyang, Jeffrey Wu, Xu~Jiang, Diogo Almeida, Carroll Wainwright, Pamela Mishkin, Chong Zhang, Sandhini Agarwal, Katarina Slama, Alex Ray, John Schulman, Jacob Hilton, Fraser Kelton, Luke Miller, Maddie Simens, Amanda Askell, Peter Welinder, Paul~F Christiano, Jan Leike, and Ryan Lowe.
\newblock Training language models to follow instructions with human feedback.
\newblock In S.~Koyejo, S.~Mohamed, A.~Agarwal, D.~Belgrave, K.~Cho, and A.~Oh (eds.), \emph{Advances in Neural Information Processing Systems}, volume~35, pp.\  27730--27744. Curran Associates, Inc., 2022.
\newblock URL \url{https://proceedings.neurips.cc/paper_files/paper/2022/file/b1efde53be364a73914f58805a001731-Paper-Conference.pdf}.

\bibitem[Sellam et~al.(2020)Sellam, Das, and Parikh]{bleurt}
Thibault Sellam, Dipanjan Das, and Ankur~P Parikh.
\newblock Bleurt: Learning robust metrics for text generation.
\newblock In \emph{Proceedings of ACL}, 2020.

\bibitem[Team et~al.(2023)Team, Anil, Borgeaud, Wu, Alayrac, Yu, Soricut, Schalkwyk, Dai, Hauth, Millican, Silver, Petrov, Johnson, Antonoglou, Schrittwieser, Glaese, Chen, Pitler, Lillicrap, Lazaridou, Firat, Molloy, Isard, Barham, Hennigan, Lee, Viola, Reynolds, Xu, Doherty, Collins, Meyer, Rutherford, Moreira, Ayoub, Goel, Tucker, Piqueras, Krikun, Barr, Savinov, Danihelka, Roelofs, White, Andreassen, von Glehn, Yagati, Kazemi, Gonzalez, Khalman, Sygnowski, Frechette, Smith, Culp, Proleev, Luan, Chen, Lottes, Schucher, Lebron, Rrustemi, Clay, Crone, Kocisky, Zhao, Perz, Yu, Howard, Bloniarz, Rae, Lu, Sifre, Maggioni, Alcober, Garrette, Barnes, Thakoor, Austin, Barth-Maron, Wong, Joshi, Chaabouni, Fatiha, Ahuja, Liu, Li, Cogan, Chen, Jia, Gu, Zhang, Grimstad, Hartman, Chadwick, Tomar, Garcia, Senter, Taropa, Pillai, Devlin, Laskin, de~Las~Casas, Valter, Tao, Blanco, Badia, Reitter, Chen, Brennan, Rivera, Brin, Iqbal, Surita, Labanowski, Rao, Winkler, Parisotto, Gu, Olszewska, Zhang, Addanki, Miech, Louis,
  Shafey, Teplyashin, Brown, Catt, Attaluri, Balaguer, Xiang, Wang, Ashwood, Briukhov, Webson, Ganapathy, Sanghavi, Kannan, Chang, Stjerngren, Djolonga, Sun, Bapna, Aitchison, Pejman, Michalewski, Yu, Wang, Love, Ahn, Bloxwich, Han, Humphreys, Sellam, Bradbury, Godbole, Samangooei, Damoc, Kaskasoli, Arnold, Vasudevan, Agrawal, Riesa, Lepikhin, Tanburn, Srinivasan, Lim, Hodkinson, Shyam, Ferret, Hand, Garg, Paine, Li, Li, Giang, Neitz, Abbas, York, Reid, Cole, Chowdhery, Das, Rogozińska, Nikolaev, Sprechmann, Nado, Zilka, Prost, He, Monteiro, Mishra, Welty, Newlan, Jia, Allamanis, Hu, de~Liedekerke, Gilmer, Saroufim, Rijhwani, Hou, Shrivastava, Baddepudi, Goldin, Ozturel, Cassirer, Xu, Sohn, Sachan, Amplayo, Swanson, Petrova, Narayan, Guez, Brahma, Landon, Patel, Zhao, Villela, Wang, Jia, Rahtz, Giménez, Yeung, Lin, Keeling, Georgiev, Mincu, Wu, Haykal, Saputro, Vodrahalli, Qin, Cankara, Sharma, Fernando, Hawkins, Neyshabur, Kim, Hutter, Agrawal, Castro-Ros, van~den Driessche, Wang, Yang, yiin Chang,
  Komarek, McIlroy, Lučić, Zhang, Farhan, Sharman, Natsev, Michel, Cheng, Bansal, Qiao, Cao, Shakeri, Butterfield, Chung, Rubenstein, Agrawal, Mensch, Soparkar, Lenc, Chung, Pope, Maggiore, Kay, Jhakra, Wang, Maynez, Phuong, Tobin, Tacchetti, Trebacz, Robinson, Katariya, Riedel, Bailey, Xiao, Ghelani, Aroyo, Slone, Houlsby, Xiong, Yang, Gribovskaya, Adler, Wirth, Lee, Li, Kagohara, Pavagadhi, Bridgers, Bortsova, Ghemawat, Ahmed, Liu, Powell, Bolina, Iinuma, Zablotskaia, Besley, Chung, Dozat, Comanescu, Si, Greer, Su, Polacek, Kaufman, Tokumine, Hu, Buchatskaya, Miao, Elhawaty, Siddhant, Tomasev, Xing, Greer, Miller, Ashraf, Roy, Zhang, Ma, Filos, Besta, Blevins, Klimenko, Yeh, Changpinyo, Mu, Chang, Pajarskas, Muir, Cohen, Lan, Haridasan, Marathe, Hansen, Douglas, Samuel, Wang, Austin, Lan, Jiang, Chiu, Lorenzo, Sjösund, Cevey, Gleicher, Avrahami, Boral, Srinivasan, Selo, May, Aisopos, Hussenot, Soares, Baumli, Chang, Recasens, Caine, Pritzel, Pavetic, Pardo, Gergely, Frye, Ramasesh, Horgan, Badola,
  Kassner, Roy, Dyer, Campos, Tomala, Tang, Badawy, White, Mustafa, Lang, Jindal, Vikram, Gong, Caelles, Hemsley, Thornton, Feng, Stokowiec, Zheng, Thacker, Çağlar Ünlü, Zhang, Saleh, Svensson, Bileschi, Patil, Anand, Ring, Tsihlas, Vezer, Selvi, Shevlane, Rodriguez, Kwiatkowski, Daruki, Rong, Dafoe, FitzGerald, Gu-Lemberg, Khan, Hendricks, Pellat, Feinberg, Cobon-Kerr, Sainath, Rauh, Hashemi, Ives, Hasson, Li, Noland, Cao, Byrd, Hou, Wang, Sottiaux, Paganini, Lespiau, Moufarek, Hassan, Shivakumar, van Amersfoort, Mandhane, Joshi, Goyal, Tung, Brock, Sheahan, Misra, Li, Rakićević, Dehghani, Liu, Mittal, Oh, Noury, Sezener, Huot, Lamm, Cao, Chen, Elsayed, Chi, Mahdieh, Tenney, Hua, Petrychenko, Kane, Scandinaro, Jain, Uesato, Datta, Sadovsky, Bunyan, Rabiej, Wu, Zhang, Vasudevan, Leurent, Alnahlawi, Georgescu, Wei, Zheng, Chan, Rabinovitch, Stanczyk, Zhang, Steiner, Naskar, Azzam, Johnson, Paszke, Chiu, Elias, Mohiuddin, Muhammad, Miao, Lee, Vieillard, Potluri, Park, Davoodi, Zhang, Stanway, Garmon,
  Karmarkar, Dong, Lee, Kumar, Zhou, Evens, Isaac, Chen, Jia, Levskaya, Zhu, Gorgolewski, Grabowski, Mao, Magni, Yao, Snaider, Casagrande, Suganthan, Palmer, Irving, Loper, Faruqui, Arkatkar, Chen, Shafran, Fink, Castaño, Giannoumis, Kim, Rybiński, Sreevatsa, Prendki, Soergel, Goedeckemeyer, Gierke, Jafari, Gaba, Wiesner, Wright, Wei, Vashisht, Kulizhskaya, Hoover, Le, Li, Iwuanyanwu, Liu, Ramirez, Khorlin, Cui, LIN, Georgiev, Wu, Aguilar, Pallo, Chakladar, Repina, Wu, van~der Weide, Ponnapalli, Kaplan, Simsa, Li, Dousse, Yang, Piper, Ie, Lui, Pasumarthi, Lintz, Vijayakumar, Thiet, Andor, Valenzuela, Paduraru, Peng, Lee, Zhang, Greene, Nguyen, Kurylowicz, Velury, Krause, Hardin, Dixon, Janzer, Choo, Feng, Zhang, Singhal, Latkar, Zhang, Le, Abellan, Du, McKinnon, Antropova, Bolukbasi, Keller, Reid, Finchelstein, Raad, Crocker, Hawkins, Dadashi, Gaffney, Lall, Franko, Filonov, Bulanova, Leblond, Yadav, Chung, Askham, Cobo, Xu, Fischer, Xu, Sorokin, Alberti, Lin, Evans, Zhou, Dimitriev, Forbes, Banarse, Tung,
  Liu, Omernick, Bishop, Kumar, Sterneck, Foley, Jain, Mishra, Xia, Bos, Cideron, Amid, Piccinno, Wang, Banzal, Gurita, Noga, Shah, Mankowitz, Polozov, Kushman, Krakovna, Brown, Bateni, Duan, Firoiu, Thotakuri, Natan, Mohananey, Geist, Mudgal, Girgin, Li, Ye, Roval, Tojo, Kwong, Lee-Thorp, Yew, Yuan, Bagri, Sinopalnikov, Ramos, Mellor, Sharma, Severyn, Lai, Wu, Cheng, Miller, Sonnerat, Vnukov, Greig, Beattie, Caveness, Bai, Eisenschlos, Korchemniy, Tsai, Jasarevic, Kong, Dao, Zheng, Liu, Yang, Zhu, Geller, Teh, Sanmiya, Gladchenko, Trdin, Sozanschi, Toyama, Rosen, Tavakkol, Xue, Elkind, Woodman, Carpenter, Papamakarios, Kemp, Kafle, Grunina, Sinha, Talbert, Goyal, Wu, Owusu-Afriyie, Du, Thornton, Pont-Tuset, Narayana, Li, Fatehi, Wieting, Ajmeri, Uria, Zhu, Ko, Knight, Héliou, Niu, Gu, Pang, Tran, Li, Levine, Stolovich, Kalb, Santamaria-Fernandez, Goenka, Yustalim, Strudel, Elqursh, Lakshminarayanan, Deck, Upadhyay, Lee, Dusenberry, Li, Wang, Levin, Hoffmann, Holtmann-Rice, Bachem, Yue, Arora, Malmi,
  Mirylenka, Tan, Koh, Yeganeh, Põder, Zheng, Pongetti, Tariq, Sun, Ionita, Seyedhosseini, Tafti, Kotikalapudi, Liu, Gulati, Liu, Ye, Chrzaszcz, Wang, Sethi, Li, Brown, Singh, Fan, Parisi, Stanton, Kuang, Koverkathu, Choquette-Choo, Li, Lu, Ittycheriah, Shroff, Sun, Varadarajan, Bahargam, Willoughby, Gaddy, Dasgupta, Desjardins, Cornero, Robenek, Mittal, Albrecht, Shenoy, Moiseev, Jacobsson, Ghaffarkhah, Rivière, Walton, Crepy, Parrish, Liu, Zhou, Farabet, Radebaugh, Srinivasan, van~der Salm, Fidjeland, Scellato, Latorre-Chimoto, Klimczak-Plucińska, Bridson, de~Cesare, Hudson, Mendolicchio, Walker, Morris, Penchev, Mauger, Guseynov, Reid, Odoom, Loher, Cotruta, Yenugula, Grewe, Petrushkina, Duerig, Sanchez, Yadlowsky, Shen, Globerson, Kurzrok, Webb, Dua, Li, Lahoti, Bhupatiraju, Hurt, Qureshi, Agarwal, Shani, Eyal, Khare, Belle, Wang, Tekur, Kale, Wei, Sang, Saeta, Liechty, Sun, Zhao, Lee, Nayak, Fritz, Vuyyuru, Aslanides, Vyas, Wicke, Ma, Bilal, Eltyshev, Balle, Martin, Cate, Manyika, Amiri, Kim, Xiong,
  Kang, Luisier, Tripuraneni, Madras, Guo, Waters, Wang, Ainslie, Baldridge, Zhang, Pruthi, Bauer, Yang, Mansour, Gelman, Xu, Polovets, Liu, Cai, Chen, Sheng, Xue, Ozair, Yu, Angermueller, Li, Wang, Wiesinger, Koukoumidis, Tian, Iyer, Gurumurthy, Goldenson, Shah, Blake, Yu, Urbanowicz, Palomaki, Fernando, Brooks, Durden, Mehta, Momchev, Rahimtoroghi, Georgaki, Raul, Ruder, Redshaw, Lee, Jalan, Li, Perng, Hechtman, Schuh, Nasr, Chen, Milan, Mikulik, Strohman, Franco, Green, Hassabis, Kavukcuoglu, Dean, and Vinyals]{gemini}
Gemini Team, Rohan Anil, Sebastian Borgeaud, Yonghui Wu, Jean-Baptiste Alayrac, Jiahui Yu, Radu Soricut, Johan Schalkwyk, Andrew~M. Dai, Anja Hauth, Katie Millican, David Silver, Slav Petrov, Melvin Johnson, Ioannis Antonoglou, Julian Schrittwieser, Amelia Glaese, Jilin Chen, Emily Pitler, Timothy Lillicrap, Angeliki Lazaridou, Orhan Firat, James Molloy, Michael Isard, Paul~R. Barham, Tom Hennigan, Benjamin Lee, Fabio Viola, Malcolm Reynolds, Yuanzhong Xu, Ryan Doherty, Eli Collins, Clemens Meyer, Eliza Rutherford, Erica Moreira, Kareem Ayoub, Megha Goel, George Tucker, Enrique Piqueras, Maxim Krikun, Iain Barr, Nikolay Savinov, Ivo Danihelka, Becca Roelofs, Anaïs White, Anders Andreassen, Tamara von Glehn, Lakshman Yagati, Mehran Kazemi, Lucas Gonzalez, Misha Khalman, Jakub Sygnowski, Alexandre Frechette, Charlotte Smith, Laura Culp, Lev Proleev, Yi~Luan, Xi~Chen, James Lottes, Nathan Schucher, Federico Lebron, Alban Rrustemi, Natalie Clay, Phil Crone, Tomas Kocisky, Jeffrey Zhao, Bartek Perz, Dian Yu,
  Heidi Howard, Adam Bloniarz, Jack~W. Rae, Han Lu, Laurent Sifre, Marcello Maggioni, Fred Alcober, Dan Garrette, Megan Barnes, Shantanu Thakoor, Jacob Austin, Gabriel Barth-Maron, William Wong, Rishabh Joshi, Rahma Chaabouni, Deeni Fatiha, Arun Ahuja, Ruibo Liu, Yunxuan Li, Sarah Cogan, Jeremy Chen, Chao Jia, Chenjie Gu, Qiao Zhang, Jordan Grimstad, Ale~Jakse Hartman, Martin Chadwick, Gaurav~Singh Tomar, Xavier Garcia, Evan Senter, Emanuel Taropa, Thanumalayan~Sankaranarayana Pillai, Jacob Devlin, Michael Laskin, Diego de~Las~Casas, Dasha Valter, Connie Tao, Lorenzo Blanco, Adrià~Puigdomènech Badia, David Reitter, Mianna Chen, Jenny Brennan, Clara Rivera, Sergey Brin, Shariq Iqbal, Gabriela Surita, Jane Labanowski, Abhi Rao, Stephanie Winkler, Emilio Parisotto, Yiming Gu, Kate Olszewska, Yujing Zhang, Ravi Addanki, Antoine Miech, Annie Louis, Laurent~El Shafey, Denis Teplyashin, Geoff Brown, Elliot Catt, Nithya Attaluri, Jan Balaguer, Jackie Xiang, Pidong Wang, Zoe Ashwood, Anton Briukhov, Albert Webson,
  Sanjay Ganapathy, Smit Sanghavi, Ajay Kannan, Ming-Wei Chang, Axel Stjerngren, Josip Djolonga, Yuting Sun, Ankur Bapna, Matthew Aitchison, Pedram Pejman, Henryk Michalewski, Tianhe Yu, Cindy Wang, Juliette Love, Junwhan Ahn, Dawn Bloxwich, Kehang Han, Peter Humphreys, Thibault Sellam, James Bradbury, Varun Godbole, Sina Samangooei, Bogdan Damoc, Alex Kaskasoli, Sébastien M.~R. Arnold, Vijay Vasudevan, Shubham Agrawal, Jason Riesa, Dmitry Lepikhin, Richard Tanburn, Srivatsan Srinivasan, Hyeontaek Lim, Sarah Hodkinson, Pranav Shyam, Johan Ferret, Steven Hand, Ankush Garg, Tom~Le Paine, Jian Li, Yujia Li, Minh Giang, Alexander Neitz, Zaheer Abbas, Sarah York, Machel Reid, Elizabeth Cole, Aakanksha Chowdhery, Dipanjan Das, Dominika Rogozińska, Vitaly Nikolaev, Pablo Sprechmann, Zachary Nado, Lukas Zilka, Flavien Prost, Luheng He, Marianne Monteiro, Gaurav Mishra, Chris Welty, Josh Newlan, Dawei Jia, Miltiadis Allamanis, Clara~Huiyi Hu, Raoul de~Liedekerke, Justin Gilmer, Carl Saroufim, Shruti Rijhwani, Shaobo
  Hou, Disha Shrivastava, Anirudh Baddepudi, Alex Goldin, Adnan Ozturel, Albin Cassirer, Yunhan Xu, Daniel Sohn, Devendra Sachan, Reinald~Kim Amplayo, Craig Swanson, Dessie Petrova, Shashi Narayan, Arthur Guez, Siddhartha Brahma, Jessica Landon, Miteyan Patel, Ruizhe Zhao, Kevin Villela, Luyu Wang, Wenhao Jia, Matthew Rahtz, Mai Giménez, Legg Yeung, Hanzhao Lin, James Keeling, Petko Georgiev, Diana Mincu, Boxi Wu, Salem Haykal, Rachel Saputro, Kiran Vodrahalli, James Qin, Zeynep Cankara, Abhanshu Sharma, Nick Fernando, Will Hawkins, Behnam Neyshabur, Solomon Kim, Adrian Hutter, Priyanka Agrawal, Alex Castro-Ros, George van~den Driessche, Tao Wang, Fan Yang, Shuo yiin Chang, Paul Komarek, Ross McIlroy, Mario Lučić, Guodong Zhang, Wael Farhan, Michael Sharman, Paul Natsev, Paul Michel, Yong Cheng, Yamini Bansal, Siyuan Qiao, Kris Cao, Siamak Shakeri, Christina Butterfield, Justin Chung, Paul~Kishan Rubenstein, Shivani Agrawal, Arthur Mensch, Kedar Soparkar, Karel Lenc, Timothy Chung, Aedan Pope, Loren
  Maggiore, Jackie Kay, Priya Jhakra, Shibo Wang, Joshua Maynez, Mary Phuong, Taylor Tobin, Andrea Tacchetti, Maja Trebacz, Kevin Robinson, Yash Katariya, Sebastian Riedel, Paige Bailey, Kefan Xiao, Nimesh Ghelani, Lora Aroyo, Ambrose Slone, Neil Houlsby, Xuehan Xiong, Zhen Yang, Elena Gribovskaya, Jonas Adler, Mateo Wirth, Lisa Lee, Music Li, Thais Kagohara, Jay Pavagadhi, Sophie Bridgers, Anna Bortsova, Sanjay Ghemawat, Zafarali Ahmed, Tianqi Liu, Richard Powell, Vijay Bolina, Mariko Iinuma, Polina Zablotskaia, James Besley, Da-Woon Chung, Timothy Dozat, Ramona Comanescu, Xiance Si, Jeremy Greer, Guolong Su, Martin Polacek, Raphaël~Lopez Kaufman, Simon Tokumine, Hexiang Hu, Elena Buchatskaya, Yingjie Miao, Mohamed Elhawaty, Aditya Siddhant, Nenad Tomasev, Jinwei Xing, Christina Greer, Helen Miller, Shereen Ashraf, Aurko Roy, Zizhao Zhang, Ada Ma, Angelos Filos, Milos Besta, Rory Blevins, Ted Klimenko, Chih-Kuan Yeh, Soravit Changpinyo, Jiaqi Mu, Oscar Chang, Mantas Pajarskas, Carrie Muir, Vered Cohen,
  Charline~Le Lan, Krishna Haridasan, Amit Marathe, Steven Hansen, Sholto Douglas, Rajkumar Samuel, Mingqiu Wang, Sophia Austin, Chang Lan, Jiepu Jiang, Justin Chiu, Jaime~Alonso Lorenzo, Lars~Lowe Sjösund, Sébastien Cevey, Zach Gleicher, Thi Avrahami, Anudhyan Boral, Hansa Srinivasan, Vittorio Selo, Rhys May, Konstantinos Aisopos, Léonard Hussenot, Livio~Baldini Soares, Kate Baumli, Michael~B. Chang, Adrià Recasens, Ben Caine, Alexander Pritzel, Filip Pavetic, Fabio Pardo, Anita Gergely, Justin Frye, Vinay Ramasesh, Dan Horgan, Kartikeya Badola, Nora Kassner, Subhrajit Roy, Ethan Dyer, Víctor Campos, Alex Tomala, Yunhao Tang, Dalia~El Badawy, Elspeth White, Basil Mustafa, Oran Lang, Abhishek Jindal, Sharad Vikram, Zhitao Gong, Sergi Caelles, Ross Hemsley, Gregory Thornton, Fangxiaoyu Feng, Wojciech Stokowiec, Ce~Zheng, Phoebe Thacker, Çağlar Ünlü, Zhishuai Zhang, Mohammad Saleh, James Svensson, Max Bileschi, Piyush Patil, Ankesh Anand, Roman Ring, Katerina Tsihlas, Arpi Vezer, Marco Selvi, Toby
  Shevlane, Mikel Rodriguez, Tom Kwiatkowski, Samira Daruki, Keran Rong, Allan Dafoe, Nicholas FitzGerald, Keren Gu-Lemberg, Mina Khan, Lisa~Anne Hendricks, Marie Pellat, Vladimir Feinberg, James Cobon-Kerr, Tara Sainath, Maribeth Rauh, Sayed~Hadi Hashemi, Richard Ives, Yana Hasson, YaGuang Li, Eric Noland, Yuan Cao, Nathan Byrd, Le~Hou, Qingze Wang, Thibault Sottiaux, Michela Paganini, Jean-Baptiste Lespiau, Alexandre Moufarek, Samer Hassan, Kaushik Shivakumar, Joost van Amersfoort, Amol Mandhane, Pratik Joshi, Anirudh Goyal, Matthew Tung, Andrew Brock, Hannah Sheahan, Vedant Misra, Cheng Li, Nemanja Rakićević, Mostafa Dehghani, Fangyu Liu, Sid Mittal, Junhyuk Oh, Seb Noury, Eren Sezener, Fantine Huot, Matthew Lamm, Nicola~De Cao, Charlie Chen, Gamaleldin Elsayed, Ed~Chi, Mahdis Mahdieh, Ian Tenney, Nan Hua, Ivan Petrychenko, Patrick Kane, Dylan Scandinaro, Rishub Jain, Jonathan Uesato, Romina Datta, Adam Sadovsky, Oskar Bunyan, Dominik Rabiej, Shimu Wu, John Zhang, Gautam Vasudevan, Edouard Leurent,
  Mahmoud Alnahlawi, Ionut Georgescu, Nan Wei, Ivy Zheng, Betty Chan, Pam~G Rabinovitch, Piotr Stanczyk, Ye~Zhang, David Steiner, Subhajit Naskar, Michael Azzam, Matthew Johnson, Adam Paszke, Chung-Cheng Chiu, Jaume~Sanchez Elias, Afroz Mohiuddin, Faizan Muhammad, Jin Miao, Andrew Lee, Nino Vieillard, Sahitya Potluri, Jane Park, Elnaz Davoodi, Jiageng Zhang, Jeff Stanway, Drew Garmon, Abhijit Karmarkar, Zhe Dong, Jong Lee, Aviral Kumar, Luowei Zhou, Jonathan Evens, William Isaac, Zhe Chen, Johnson Jia, Anselm Levskaya, Zhenkai Zhu, Chris Gorgolewski, Peter Grabowski, Yu~Mao, Alberto Magni, Kaisheng Yao, Javier Snaider, Norman Casagrande, Paul Suganthan, Evan Palmer, Geoffrey Irving, Edward Loper, Manaal Faruqui, Isha Arkatkar, Nanxin Chen, Izhak Shafran, Michael Fink, Alfonso Castaño, Irene Giannoumis, Wooyeol Kim, Mikołaj Rybiński, Ashwin Sreevatsa, Jennifer Prendki, David Soergel, Adrian Goedeckemeyer, Willi Gierke, Mohsen Jafari, Meenu Gaba, Jeremy Wiesner, Diana~Gage Wright, Yawen Wei, Harsha Vashisht,
  Yana Kulizhskaya, Jay Hoover, Maigo Le, Lu~Li, Chimezie Iwuanyanwu, Lu~Liu, Kevin Ramirez, Andrey Khorlin, Albert Cui, Tian LIN, Marin Georgiev, Marcus Wu, Ricardo Aguilar, Keith Pallo, Abhishek Chakladar, Alena Repina, Xihui Wu, Tom van~der Weide, Priya Ponnapalli, Caroline Kaplan, Jiri Simsa, Shuangfeng Li, Olivier Dousse, Fan Yang, Jeff Piper, Nathan Ie, Minnie Lui, Rama Pasumarthi, Nathan Lintz, Anitha Vijayakumar, Lam~Nguyen Thiet, Daniel Andor, Pedro Valenzuela, Cosmin Paduraru, Daiyi Peng, Katherine Lee, Shuyuan Zhang, Somer Greene, Duc~Dung Nguyen, Paula Kurylowicz, Sarmishta Velury, Sebastian Krause, Cassidy Hardin, Lucas Dixon, Lili Janzer, Kiam Choo, Ziqiang Feng, Biao Zhang, Achintya Singhal, Tejasi Latkar, Mingyang Zhang, Quoc Le, Elena~Allica Abellan, Dayou Du, Dan McKinnon, Natasha Antropova, Tolga Bolukbasi, Orgad Keller, David Reid, Daniel Finchelstein, Maria~Abi Raad, Remi Crocker, Peter Hawkins, Robert Dadashi, Colin Gaffney, Sid Lall, Ken Franko, Egor Filonov, Anna Bulanova, Rémi
  Leblond, Vikas Yadav, Shirley Chung, Harry Askham, Luis~C. Cobo, Kelvin Xu, Felix Fischer, Jun Xu, Christina Sorokin, Chris Alberti, Chu-Cheng Lin, Colin Evans, Hao Zhou, Alek Dimitriev, Hannah Forbes, Dylan Banarse, Zora Tung, Jeremiah Liu, Mark Omernick, Colton Bishop, Chintu Kumar, Rachel Sterneck, Ryan Foley, Rohan Jain, Swaroop Mishra, Jiawei Xia, Taylor Bos, Geoffrey Cideron, Ehsan Amid, Francesco Piccinno, Xingyu Wang, Praseem Banzal, Petru Gurita, Hila Noga, Premal Shah, Daniel~J. Mankowitz, Alex Polozov, Nate Kushman, Victoria Krakovna, Sasha Brown, MohammadHossein Bateni, Dennis Duan, Vlad Firoiu, Meghana Thotakuri, Tom Natan, Anhad Mohananey, Matthieu Geist, Sidharth Mudgal, Sertan Girgin, Hui Li, Jiayu Ye, Ofir Roval, Reiko Tojo, Michael Kwong, James Lee-Thorp, Christopher Yew, Quan Yuan, Sumit Bagri, Danila Sinopalnikov, Sabela Ramos, John Mellor, Abhishek Sharma, Aliaksei Severyn, Jonathan Lai, Kathy Wu, Heng-Tze Cheng, David Miller, Nicolas Sonnerat, Denis Vnukov, Rory Greig, Jennifer
  Beattie, Emily Caveness, Libin Bai, Julian Eisenschlos, Alex Korchemniy, Tomy Tsai, Mimi Jasarevic, Weize Kong, Phuong Dao, Zeyu Zheng, Frederick Liu, Fan Yang, Rui Zhu, Mark Geller, Tian~Huey Teh, Jason Sanmiya, Evgeny Gladchenko, Nejc Trdin, Andrei Sozanschi, Daniel Toyama, Evan Rosen, Sasan Tavakkol, Linting Xue, Chen Elkind, Oliver Woodman, John Carpenter, George Papamakarios, Rupert Kemp, Sushant Kafle, Tanya Grunina, Rishika Sinha, Alice Talbert, Abhimanyu Goyal, Diane Wu, Denese Owusu-Afriyie, Cosmo Du, Chloe Thornton, Jordi Pont-Tuset, Pradyumna Narayana, Jing Li, Sabaer Fatehi, John Wieting, Omar Ajmeri, Benigno Uria, Tao Zhu, Yeongil Ko, Laura Knight, Amélie Héliou, Ning Niu, Shane Gu, Chenxi Pang, Dustin Tran, Yeqing Li, Nir Levine, Ariel Stolovich, Norbert Kalb, Rebeca Santamaria-Fernandez, Sonam Goenka, Wenny Yustalim, Robin Strudel, Ali Elqursh, Balaji Lakshminarayanan, Charlie Deck, Shyam Upadhyay, Hyo Lee, Mike Dusenberry, Zonglin Li, Xuezhi Wang, Kyle Levin, Raphael Hoffmann, Dan
  Holtmann-Rice, Olivier Bachem, Summer Yue, Sho Arora, Eric Malmi, Daniil Mirylenka, Qijun Tan, Christy Koh, Soheil~Hassas Yeganeh, Siim Põder, Steven Zheng, Francesco Pongetti, Mukarram Tariq, Yanhua Sun, Lucian Ionita, Mojtaba Seyedhosseini, Pouya Tafti, Ragha Kotikalapudi, Zhiyu Liu, Anmol Gulati, Jasmine Liu, Xinyu Ye, Bart Chrzaszcz, Lily Wang, Nikhil Sethi, Tianrun Li, Ben Brown, Shreya Singh, Wei Fan, Aaron Parisi, Joe Stanton, Chenkai Kuang, Vinod Koverkathu, Christopher~A. Choquette-Choo, Yunjie Li, TJ~Lu, Abe Ittycheriah, Prakash Shroff, Pei Sun, Mani Varadarajan, Sanaz Bahargam, Rob Willoughby, David Gaddy, Ishita Dasgupta, Guillaume Desjardins, Marco Cornero, Brona Robenek, Bhavishya Mittal, Ben Albrecht, Ashish Shenoy, Fedor Moiseev, Henrik Jacobsson, Alireza Ghaffarkhah, Morgane Rivière, Alanna Walton, Clément Crepy, Alicia Parrish, Yuan Liu, Zongwei Zhou, Clement Farabet, Carey Radebaugh, Praveen Srinivasan, Claudia van~der Salm, Andreas Fidjeland, Salvatore Scellato, Eri Latorre-Chimoto,
  Hanna Klimczak-Plucińska, David Bridson, Dario de~Cesare, Tom Hudson, Piermaria Mendolicchio, Lexi Walker, Alex Morris, Ivo Penchev, Matthew Mauger, Alexey Guseynov, Alison Reid, Seth Odoom, Lucia Loher, Victor Cotruta, Madhavi Yenugula, Dominik Grewe, Anastasia Petrushkina, Tom Duerig, Antonio Sanchez, Steve Yadlowsky, Amy Shen, Amir Globerson, Adam Kurzrok, Lynette Webb, Sahil Dua, Dong Li, Preethi Lahoti, Surya Bhupatiraju, Dan Hurt, Haroon Qureshi, Ananth Agarwal, Tomer Shani, Matan Eyal, Anuj Khare, Shreyas~Rammohan Belle, Lei Wang, Chetan Tekur, Mihir~Sanjay Kale, Jinliang Wei, Ruoxin Sang, Brennan Saeta, Tyler Liechty, Yi~Sun, Yao Zhao, Stephan Lee, Pandu Nayak, Doug Fritz, Manish~Reddy Vuyyuru, John Aslanides, Nidhi Vyas, Martin Wicke, Xiao Ma, Taylan Bilal, Evgenii Eltyshev, Daniel Balle, Nina Martin, Hardie Cate, James Manyika, Keyvan Amiri, Yelin Kim, Xi~Xiong, Kai Kang, Florian Luisier, Nilesh Tripuraneni, David Madras, Mandy Guo, Austin Waters, Oliver Wang, Joshua Ainslie, Jason Baldridge, Han
  Zhang, Garima Pruthi, Jakob Bauer, Feng Yang, Riham Mansour, Jason Gelman, Yang Xu, George Polovets, Ji~Liu, Honglong Cai, Warren Chen, XiangHai Sheng, Emily Xue, Sherjil Ozair, Adams Yu, Christof Angermueller, Xiaowei Li, Weiren Wang, Julia Wiesinger, Emmanouil Koukoumidis, Yuan Tian, Anand Iyer, Madhu Gurumurthy, Mark Goldenson, Parashar Shah, MK~Blake, Hongkun Yu, Anthony Urbanowicz, Jennimaria Palomaki, Chrisantha Fernando, Kevin Brooks, Ken Durden, Harsh Mehta, Nikola Momchev, Elahe Rahimtoroghi, Maria Georgaki, Amit Raul, Sebastian Ruder, Morgan Redshaw, Jinhyuk Lee, Komal Jalan, Dinghua Li, Ginger Perng, Blake Hechtman, Parker Schuh, Milad Nasr, Mia Chen, Kieran Milan, Vladimir Mikulik, Trevor Strohman, Juliana Franco, Tim Green, Demis Hassabis, Koray Kavukcuoglu, Jeffrey Dean, and Oriol Vinyals.
\newblock Gemini: A family of highly capable multimodal models, 2023.

\bibitem[Vu et~al.(2023)Vu, Iyyer, Wang, Constant, Wei, Wei, Tar, Sung, Zhou, Le, et~al.]{vu2023freshllms}
Tu~Vu, Mohit Iyyer, Xuezhi Wang, Noah Constant, Jerry Wei, Jason Wei, Chris Tar, Yun-Hsuan Sung, Denny Zhou, Quoc Le, et~al.
\newblock Freshllms: Refreshing large language models with search engine augmentation.
\newblock \emph{arXiv preprint arXiv:2310.03214}, 2023.

\bibitem[Wang et~al.(2023)Wang, Feng, Wang, Shi, Balachandran, He, and Tsvetkov]{wang2023resolving}
Yike Wang, Shangbin Feng, Heng Wang, Weijia Shi, Vidhisha Balachandran, Tianxing He, and Yulia Tsvetkov.
\newblock Resolving knowledge conflicts in large language models.
\newblock \emph{arXiv preprint arXiv:2310.00935}, 2023.

\bibitem[Wei et~al.(2019)Wei, Liu, Zhu, Zang, Zhou, Han, and Hu]{wei2019esa}
Dongjun Wei, Yaxin Liu, Fuqing Zhu, Liangjun Zang, Wei Zhou, Jizhong Han, and Songlin Hu.
\newblock Esa: entity summarization with attention.
\newblock \emph{arXiv preprint arXiv:1905.10625}, 2019.

\bibitem[Yang et~al.(2021)Yang, Hao, Yang, De~Maio, Nasridinov, Min, and Yang]{yang2021incremental}
Erhe Yang, Fei Hao, Yixuan Yang, Carmen De~Maio, Aziz Nasridinov, Geyong Min, and Laurence~T Yang.
\newblock Incremental entity summarization with formal concept analysis.
\newblock \emph{IEEE Transactions on Services Computing}, 15\penalty0 (6):\penalty0 3289--3303, 2021.

\end{thebibliography}
\bibliographystyle{colm2024_conference}

\appendix
\section{Appendix}
\subsection{Limitations}
\label{app:limitation}
Firstly, although the evaluation uses three LLMs (including Gemini and GPT-3.5), incorporating additional open-source models would strengthen the findings. Secondly, the chosen LLM-based evaluation metrics can be computationally expensive and time-consuming to execute. Finally, while the automated critique and revision module demonstrates high accuracy (96\%) on a sample set, reviewing the entire dataset with human annotators could potentially achieve complete alignment (100\%) between generated summary tables and paragraphs.

\subsection{Detailed Dataset Stats}
\label{app:dataset-stats}
We present our dataset statistics for the entity level and paragraph level in Table \ref{tab:entity-level-stats} and \ref{tab:para-level-stats}. Table \ref{tab:entity-level-stats} details the average number of attributes and values associated with individual entities. It also shows the average number of unique attributes and values observed across all entities, considering all paragraphs associated with each entity. Table \ref{tab:para-level-stats} shows the average number of ``same'', ``conflict'', ``new'' attribute and value pairs, and an average number of sentences and distractor sentences in each paragraph. 
\begin{table}[h]
\small
\centering
\begin{tabular}{l|rrrrr}
\toprule
Categories & \# attr & \# val & \# diff attr & \# diff val\\
\midrule
% \hline
% Bath \& Body Products & 21.60 & 40.70 & 13.44 & 39.06 \\
% Bedding \& Bed Linens & 22.50 & 37.80 & 13.78 & 36.17 \\
% Books \& Literature & 21.80 & 41.20 & 16.00 & 40.50 \\
% Computer \& Video Games & 22.70 & 39.90 & 14.17 & 37.72 \\
% Computers \& Electronics & 22.60 & 41.90 & 14.39 & 40.61 \\
% Drugs \& Medications & 19.70 & 38.20 & 11.94 & 37.61 \\
% Education & 23.00 & 42.60 & 15.56 & 41.94 \\
% Fashion \& Style & 23.20 & 43.50 & 15.78 & 42.78 \\
% Fruits \& Vegetables & 23.10 & 40.30 & 13.00 & 38.67 \\
% Hobbies \& Leisure & 22.90 & 44.80 & 15.44 & 43.61 \\
% Hotels \& Accommodations & 22.60 & 42.20 & 14.78 & 41.00 \\
% Household Supplies & 22.30 & 39.60 & 13.61 & 38.22 \\
% Music Technology & 22.60 & 43.50 & 14.11 & 42.39 \\
% Oral \& Dental Care & 22.50 & 43.40 & 14.56 & 42.44 \\
% Pets \& Animals & 22.50 & 37.30 & 15.22 & 34.89 \\
% Restaurants \& Bars & 22.50 & 40.40 & 14.50 & 39.11 \\
% Skin \& Nail Care & 23.80 & 40.50 & 15.72 & 39.83 \\
% Sports & 22.20 & 42.50 & 16.17 & 42.22 \\
% TV Shows \& Movies & 22.90 & 40.80 & 16.00 & 40.17 \\
% Vitamins \& Supplements & 21.10 & 38.10 & 13.28 & 37.33 \\

Bath \& Body Products & 23.70 & 44.00 & 12.06 & 43.17 \\
Bedding \& Bed Linens & 22.10 & 40.40 & 13.50 & 37.11 \\
Books \& Literature & 22.30 & 43.80 & 16.78 & 43.00 \\
Computer \& Video Games & 23.20 & 42.60 & 14.78 & 42.44 \\
Computers \& Electronics & 22.50 & 43.80 & 15.72 & 42.22 \\
Drugs \& Medications & 19.60 & 39.70 & 10.83 & 38.33 \\
Education & 23.20 & 45.20 & 14.56 & 44.50 \\
Fashion \& Style & 23.50 & 45.60 & 15.17 & 45.11 \\
Fruits \& Vegetables & 22.30 & 40.80 & 13.72 & 39.22 \\
Hobbies \& Leisure & 22.80 & 44.50 & 16.06 & 44.22 \\
Hotels \& Accommodations & 22.70 & 40.40 & 16.06 & 38.50 \\
Household Supplies & 21.60 & 40.90 & 13.50 & 38.39 \\
Music Equipment \& Technology & 21.90 & 44.30 & 13.17 & 43.06 \\
Oral \& Dental Care & 22.40 & 44.80 & 13.44 & 43.56 \\
Pets \& Animals & 22.70 & 41.70 & 15.56 & 39.72 \\
Restaurants \& Bars & 22.30 & 40.30 & 14.89 & 39.22 \\
Skin \& Nail Care & 22.90 & 40.80 & 15.94 & 39.56 \\
Sports & 22.40 & 42.70 & 16.17 & 42.61 \\
TV Shows \& Movies & 21.50 & 39.50 & 15.83 & 38.39 \\
Vitamins \& Supplements & 20.80 & 39.70 & 13.44 & 38.89 \\
\bottomrule
\end{tabular}

\caption{Entity level statistics. \# attr: average number of attributes per entity, \# val: average number of values per entity, \# diff attr: average different number of attributes across entities, \# diff val: average different number of values across entities. }
\label{tab:entity-level-stats}
\end{table}

\begin{table}[h]
\small
\centering
\begin{tabular}{l|ccccc}
\toprule
 & \# same & \# conflict & \# new &  & \\
Categories & attr-val & attr-val & attr-val & \# sent & \# dist\\
% \hline
\midrule
% Bath \& Body Products & 3.82 & 3.52 & 2.18 & 11.95 & 3.95 \\
% Bedding \& Bed Linens & 3.03 & 3.73 & 2.35 & 12.22 & 4.00 \\
% Books \& Literature & 3.48 & 3.60 & 2.15 & 11.93 & 4.00 \\
% Computer \& Video Games & 3.52 & 3.37 & 2.40 & 12.07 & 4.00 \\
% Computers \& Electronics & 3.65 & 3.38 & 2.42 & 12.60 & 4.00 \\
% Drugs \& Medications & 3.98 & 2.92 & 1.95 & 11.77 & 4.00 \\
% Education & 3.65 & 3.52 & 2.37 & 12.17 & 4.00 \\
% Fashion \& Style & 3.42 & 3.63 & 2.48 & 12.45 & 4.00 \\
% Fruits \& Vegetables & 3.77 & 3.00 & 2.47 & 12.17 & 4.00 \\
% Hobbies \& Leisure & 3.07 & 3.88 & 2.42 & 12.37 & 4.00 \\
% Hotels \& Accommodations & 3.42 & 3.58 & 2.40 & 12.03 & 3.95 \\
% Household Supplies & 3.18 & 3.73 & 2.28 & 12.17 & 4.00 \\
% Music Technology & 3.58 & 3.55 & 2.33 & 11.95 & 4.00 \\
% Oral \& Dental Care & 3.38 & 3.97 & 2.22 & 12.12 & 4.00 \\
% Pets \& Animals & 4.07 & 3.17 & 2.35 & 12.37 & 3.95 \\
% Restaurants \& Bars & 3.77 & 3.42 & 2.35 & 12.35 & 4.00 \\
% Skin \& Nail Care & 2.98 & 3.87 & 2.55 & 12.33 & 4.00 \\
% Sports & 3.75 & 3.22 & 2.45 & 12.08 & 4.00 \\
% TV Shows \& Movies & 3.55 & 3.22 & 2.55 & 11.65 & 4.00 \\
% Vitamins \& Supplements & 3.30 & 3.75 & 2.03 & 12.03 & 4.00 \\

Bath \& Body Products & 3.45 & 3.58 & 2.43 & 12.28 & 4.00 \\
Bedding \& Bed Linens & 3.50 & 3.87 & 2.23 & 12.33 & 4.00 \\
Books \& Literature & 3.68 & 3.55 & 2.30 & 11.93 & 4.00 \\
Computer \& Video Games & 3.57 & 3.35 & 2.48 & 12.13 & 4.00 \\
Computers \& Electronics & 3.65 & 3.62 & 2.33 & 12.55 & 4.00 \\
Drugs \& Medications & 4.17 & 3.22 & 1.95 & 12.12 & 4.00 \\
Education & 3.40 & 3.62 & 2.43 & 12.02 & 4.00 \\
Fashion \& Style & 3.67 & 3.52 & 2.55 & 12.70 & 4.00 \\
Fruits \& Vegetables & 3.93 & 3.02 & 2.38 & 11.88 & 4.00 \\
Hobbies \& Leisure & 3.57 & 3.52 & 2.38 & 12.72 & 4.00 \\
Hotels \& Accommodations & 3.45 & 3.67 & 2.38 & 12.33 & 4.00 \\
Household Supplies & 3.43 & 3.78 & 2.20 & 11.85 & 4.00 \\
Music Equipment \& Technology & 3.58 & 3.72 & 2.28 & 12.25 & 4.00 \\
Oral \& Dental Care & 3.40 & 3.88 & 2.20 & 11.80 & 4.00 \\
Pets \& Animals & 4.00 & 3.30 & 2.32 & 12.18 & 4.00 \\
Restaurants \& Bars & 4.02 & 3.45 & 2.18 & 12.23 & 4.00 \\
Skin \& Nail Care & 3.77 & 3.55 & 2.42 & 12.42 & 4.00 \\
Sports & 3.87 & 3.27 & 2.32 & 12.35 & 4.00 \\
TV Shows \& Movies & 3.55 & 3.42 & 2.22 & 11.97 & 4.00 \\
Vitamins \& Supplements & 3.48 & 3.63 & 2.07 & 12.08 & 4.00 \\
\bottomrule
\end{tabular}

\caption{Paragraph level statistics. \# same attr-val: average number of same attribute-value pairs between paragraphs, \# conflict attr-val: average number of conflicting attribute-value pairs between paragraphs, \# new attr-val: average number of new attribute-value pairs between paragraphs, \# sent: average number of sentences per paragraph, \# dist: average number of distracting sentences per paragraph.}
\label{tab:para-level-stats}
\end{table}

\subsection{Dataset Generation Prompts}
\label{app:dataset-generation-prompts}
Figure \ref{fig:generate-attribute-instruction} and \ref{fig:generate-entity-instruction} show prompts for generating attributes and fake entity names, respectively. Figure \ref{fig:generate-summary-instruction} presents a prompt for generating values as a summary table format. Figure \ref{fig:generate-paragraph-instruction} and \ref{fig:alignment-critique} are prompts for generating paragraphs and for aligning summary tables to paragraphs, respectively.

\begin{figure}[h]
\begin{framed}
\tt
\small
TASK: List the top 50 attributes when people summarize entity of a given category.
The attributes should be common or rare according to the request.
Attributes should be separated by '; '.
\end{framed}
\caption{Generate Attribute Instruction.}
\label{fig:generate-attribute-instruction}
\end{figure}

\begin{figure}[h]
\begin{framed}
\tt
\small
TASK: Generate 45 fake plausible entity names in the given category. \\
Make sure that entity names are unique. 
Entities should be separated by '; '.
\end{framed}
\caption{Generate Entity Name Instruction.}
\label{fig:generate-entity-instruction}
\end{figure}

\begin{figure}[h]
\begin{framed}
\tt
\small
TASK: Create a descriptive summary table for a given entity focusing on the following attributes and the given type. \\
For each attribute, generate at least three descriptive values that are: \\
 \\
1. Meaningful and informative. \\
2. Diverse in length, ranging from one word to a maximum of ten words. \\
3. Varied in style, offering a mix of user reviews, official product descriptions, and editorial insights. \\
4. Type: "Fact" should not contain any words that can be interpreted as positive or negative properties of the given entities (e.g. restrooms are well-maintained, family-friendly). \\

The summary table should have two columns: attributes and values. Ensure the values are separated by '; ' to clearly distinguish between them.
\end{framed}
\caption{Generate Default Summary Instruction.}
\label{fig:generate-summary-instruction}
\end{figure}

\begin{figure}[h]
\begin{framed}
\tt
\small
TASK: Create a paragraph for a given entity focusing on the following attributes and values. \\
For each attribute and value, generate at least one sentence that is: \\
 \\
1. Meaningful and informative, including both subjective opinions and objective facts. \\
2. Writing style should follow the given paragraph writing style. \\
3. Make sure to cite index number in summary table when generating the sentence. \\
4. Make sure to include diverse sentiments and attribute and values in the summary table. \\
5. Make sure not to change the core meaning of attribute and value pair due to writing style and sentiment. \\
 \\
The paragraph should include all index numbers, attributes, and values in the summary table.
Split sentences with a new line.
\end{framed}
\caption{Generate Paragraph Instruction.}
\label{fig:generate-paragraph-instruction}
\end{figure}

\begin{figure}[h]
\begin{framed}
\tt
\small
TASK: Verify whether the given attributes and values are described in the sentences and whether corresponding index number is cited correctly. \\
 \\
The inputs contain multiple lines, each of which starts with multiple (index, attribute, value) pairs, and a sentence can be followed or not. \\
Please output True/False for each line. \\
 \\
These are two conditions of being False: \\
1. Given attribute and value pairs do not followed by a sentence. \\
2. The context around citation number does not match with the index number in the attribute and value pairs. \\
3. Sentiment of the given attribute and value pair is incorrectly reflected in the sentence. \\
 \\
If False is outputted, please provide an explanation and revise the original sentence or generate a new sentence to describe the attribute and value for an entity and its category. \\
Revised sentence should not include any new information other than provided attribute and value. \\
Ensure that all attribute and value pairs are completely mentioned. \\
Make sure to include the index number of the attribute and value pair using square braces (e.g. [index]). \\
Do not make up any citation numbers that are not provided in (index, attribute, value) pairs. \\
The format should be as follows: "[(index1, attribute1, value1), (index2, attribute2, value2), ...];;;True;;;" or "[(index1, attribute1, value1), (index2, attribute2, value2), ...];;;False;;;Explanation;;;Revised/New sentence".
\end{framed}
\caption{Critique for Summary-Paragraph Alignment Prompt.}
\label{fig:alignment-critique}
\end{figure}

\begin{figure}[h]
\begin{framed}
\tt
\small
TASK:
Generate 10-15 complicated sentences that describe the given entity and category using the given attributes. \\
Generated sentences should: \\
 \\
1. Be meaningful and informative, including both subjective opinions and objective facts. \\
2. Be varied in style, offering a mix of user reviews, official product descriptions, and editorial insights. \\
3. Make sure to include entity name in the sentence. \\
 \\
Split sentences with a new line.
\end{framed}
\caption{Generate Irrelevant Sentence Instruction.}
\label{fig:generate-irrelevant-sentence-instruction}
\end{figure}

%------------------------ Performance across paragraph tones ------------------------
\subsection{LLM-based Evaluation Prompts}
\label{app:eval-prompts}
Figure \ref{fig:evidence-finding-prompt} shows a prompt used for LLM evidence finding (in Sec \ref{sec:eval-metrics}). In UPDATE method, we use prompts for GENERATE at 1st iteration, which are a combination of Figure \ref{fig:generate-instruction-prompt} and \ref{fig:generate-example-prompt}. Afterwards, we use prompts for UPDATE in Figure \ref{fig:update-instruction-prompt} and \ref{fig:update-example-prompt} for the subsequent iterations. Similarly, in MERGE method, we use prompts for GENERATE at 1st iteration, which are a combined version of Figure \ref{fig:generate-instruction-prompt} and \ref{fig:generate-example-prompt}. For the subsequent iterations, we employ two prompts for GENERATE (Figure \ref{fig:generate-instruction-prompt} and \ref{fig:generate-example-prompt}) MERGE (Figure \ref{fig:merge-instruction-prompt} and \ref{fig:merge-example-prompt}).

%------------------------ GENERATE Prompt ------------------------
\begin{figure}[h]
\begin{framed}
\tt
\small
Task Overview: \\
Your task involves synthesizing information from detailed descriptive paragraphs about a specific entity into a summary table. \\
This table will highlight key attributes of the entity along with their detailed descriptions derived from the given texts. \\
 \\
Instructions: \\
* Extract Descriptive Values: Focus on extracting specific, detailed information rather than general or vague adjectives like "good" or "bad." Ensure that descriptions are precise and informative. \\
* Present a Balanced View: The table should reflect a balanced perspective, including positive, negative, and neutral attributes. For attributes with mixed reviews, indicate the sources supporting each viewpoint. \\
* Attribute Selection: \\
 - Commonly Interested Attributes: Include attributes that are generally of interest for the type of entity being described. \\
 - Unique Attributes: Also identify and include unique attributes that are specifically mentioned in the provided descriptions. \\
* Citations and Evidence: Each attribute listed in the table should be supported by citations from the source paragraphs. Keep evidence concise but ensure it substantiates the listed values. \\
 \\
Structure of the Summary Table: \\
* The table should be organized into two columns: Attribute and Value. \\
* List attributes with their corresponding values, including citations indicating the source paragraph and relevant excerpts for substantiation. \\
* Citation and evidence should be paired in a [] and separated by ';'. If an attribute has multiple values, then each value should be separated by '\&\&\&'. 
\end{framed}
\caption{Instruction prompt for GENERATE.}
\label{fig:generate-instruction-prompt}
\end{figure}

\begin{figure}[h]
\begin{framed}
\tt
\small
Example: \\
Entity: San Jose Marriott Hotel \\
 \\
Paragraphs: \\
P1. Great room and service, but breakfast was lacking. We loved the spacious room and friendly staff, but the breakfast options were limited. There are two pools. \\
P2. Poor customer service overshadowed the beautiful location. The beachfront view was amazing, but dealing with unhelpful staff was frustrating. Room is comfortable. \\
P3. Exceptional dining and comfortable beds, but noisy at night. The restaurant was five-star, and the beds were very cozy, but there was a lot of street noise. \\
 \\
Summary Table: \\
| Attribute | Value | \\
| --- | --- | \\
| Room Quality | Spacious and comfortable rooms ([P1, "spacious room"]; [P2, "Room is comfortable"]) | \\
| Amenities | Two pools ([P1, "There are two pools"]) | \\
| Service | Friendly staff ([P1, "friendly staff"]) \&\&\& overshadowed by unhelpful staff ([P2, "Poor customer service overshadowed the beautiful location"]) | \\
| Location | Beautiful beachfront view ([P2, "The beachfront view was amazing"]) | \\
| Food \& Beverage | Exceptional dining experience ([P3, "Exceptional dining"]) \&\&\& limited breakfast options ([P1, "but breakfast was lacking"]) | \\
| Noise Level | Notable street noise at night ([P3, "but there was a lot of street noise"]) | \\
 \\
Your Task: \\
Generate a similar table based on the following descriptions of the specified entity. \\
Entity: < entity name > \\
 \\
Paragraphs: \\
< paragraph > \\
 \\
Proceed to generate the summary table. Output summary table format should follow the above example of Summary Table.
\end{framed}
\caption{Prompt that describes GENERATE task with one example.}
\label{fig:generate-example-prompt}
\end{figure}

%------------------------ UPDATE Prompt ------------------------
\begin{figure}[h]
\begin{framed}
\tt
\small
Task Overview: \\
You are tasked with refining and expanding an existing summary table based on new descriptive paragraphs about an entity. \\
This involves updating the table to include new information, modify existing details without removing any, and ensuring all entries are supported by evidence from the text. \\
 \\
Instructions: \\
* Update Descriptive Values: Carefully read the new paragraph(s) and identify any information that should be added to the current table entries or modify them. Focus on specific, descriptive details, avoiding vague adjectives. \\
**Do not remove any existing attributes or values**, but rather add to or revise them as necessary. \\
* Maintain a Balanced View: Ensure the updated table continues to present a balanced perspective, incorporating positive, negative, and neutral values. For any attribute with mixed evidence, update the count of sources supporting each view. \\
* Maintain a Balanced View: Ensure the updated table continues to present a balanced perspective, incorporating positive, negative, and neutral values. For any attribute with mixed evidence, update the count of sources supporting each view. All original attributes and values must be preserved in the table, with modifications only to reflect new insights or corrections based on the latest information. \\
* Attribute Revision and Addition: \\
 - Commonly Interested Attributes: Update or add attributes that are of general interest for the type of entity being described, based on the new information. \\
 - Unique Attributes: Identify and incorporate any unique attributes mentioned in the new paragraphs that were not previously included in the table. \\
* Evidence and Citations: For each updated or new attribute entry, provide citations from the new paragraphs. Strive for concise evidence that directly supports the attribute values. \\
 \\
Structure of the Updated Summary Table: \\
* Retain the two-column format: Attribute and Value. \\
* For each attribute, list the updated or new values along with citations indicating the source paragraph and relevant excerpts. Original attributes and values should remain listed, with additional information appended as necessary. \\
* Citation and evidence should be paired in a [] and separated by ';'. If an attribute has multiple values, then each value should be separated by '\&\&\&'. \\
\end{framed}
\caption{Instruction prompt for UPDATE.}
\label{fig:update-instruction-prompt}
\end{figure}

\begin{figure}[h]
\begin{framed}
\tt
\small
Example \\
Entity: San Jose Marriott Hotel \\
New Paragraph: \\
"P4. The hotel has recently renovated its lobby, which now features a modern design. Guests have also noted improvements in breakfast variety and quality." \\
 \\
Given Existing Summary Table: \\
| Attribute | Value | \\
| --- | --- | \\
| Room Quality | Spacious and comfortable rooms ([P1, "spacious room"]; [P2, "Room is comfortable"]) | \\
| Amenities | Two pools ([P1, "There are two pools"]) | \\
| Service | Friendly staff ([P1, "friendly staff"]) \&\&\& overshadowed by unhelpful staff ([P2, "Poor customer service overshadowed the beautiful location"]) | \\
 \\
Updated Summary Table: \\
| Attribute | Value | \\
| --- | --- | \\
| Room Quality | Spacious and comfortable rooms ([P1, "spacious room"]; [P2, "Room is comfortable"]) | \\
| Amenities | Two pools ([P1, "There are two pools"]) | \\
| Food \& Beverage | Exceptional dining experience ([P3, "Exceptional dining"]) \&\&\& limited breakfast options ([P1, "but breakfast was lacking"]) \&\&\& improved breakfast variety and quality ([P4, "improvements in breakfast variety and quality"])| \\
| Lobby Design | Modern design ([P4, "recently renovated its lobby, which now features a modern design"])| \\
 \\
Your Task: \\
Update the summary table with the given new descriptions of the specified entity. \\
Entity: < entity name > \\
New Paragraph: \\
< paragraph > \\
 \\
Given Existing Summary Table: \\
< existing summary table > \\
 \\
Proceed to update the summary table. Output summary table format should follow the above example of Summary Table. \\
\end{framed}
\caption{Prompt that describes UPDATE task with one example.}
\label{fig:update-example-prompt}
\end{figure}

%------------------------ MERGE Prompt ------------------------
\begin{figure}[h]
\begin{framed}
\tt
\small
Task Overview: \\
You are tasked with combining two summary tables based on existing and new descriptive paragraphs about an entity and generating an updated summary table that contains information from both tables (existing summary table and new summary table). \\
This involves updating the table to include new information, modify existing details without removing any, and ensuring all entries are supported by evidence from the text. \\
 \\
Instructions: \\
* Update Descriptive Values: Carefully read the new paragraph(s) and identify any information that should be added to the current table entries or modify them. Focus on specific, descriptive details, avoiding vague adjectives. \\
**Do not remove any existing attributes or values**, but rather add to or revise them as necessary.\\
* Maintain a Balanced View: Ensure the updated table continues to present a balanced perspective, incorporating positive, negative, and neutral values. For any attribute with mixed evidence, update the count of sources supporting each view. \\
* Maintain a Balanced View: Ensure the updated table continues to present a balanced perspective, incorporating positive, negative, and neutral values. For any attribute with mixed evidence, update the count of sources supporting each view. All original attributes and values must be preserved in the table, with modifications only to reflect new insights or corrections based on the latest information. \\
* Attribute Revision and Addition: \\
 - Commonly Interested Attributes: Update or add attributes that are of general interest for the type of entity being described, based on the new information. \\
 - Unique Attributes: Identify and incorporate any unique attributes mentioned in the new paragraphs that were not previously included in the table. \\
* Evidence and Citations: For each updated or new attribute entry, provide citations from the new paragraphs. Strive for concise evidence that directly supports the attribute values. \\
 \\
Structure of the Updated Summary Table: \\
* Retain the two-column format: Attribute and Value. \\
* For each attribute, list the updated or new values along with citations indicating the source paragraph and relevant excerpts. Original attributes and values should remain listed, with additional information appended as necessary. \\
* Citation and evidence should be paired in a [] and separated by ';'. If an attribute has multiple values, then each value should be separated by '\&\&\&'. \\
\end{framed}
\caption{Instruction prompt for MERGE.}
\label{fig:merge-instruction-prompt}
\end{figure}

%------------------------ MERGE Prompt ------------------------
\begin{figure}[h]
\begin{framed}
\tt
\small
Example \\
Entity: San Jose Marriott Hotel \\
 \\
Given Existing Summary Table: \\
| Attribute | Value | \\
| --- | --- | \\
| Room Quality | Spacious and comfortable rooms ([P1, "spacious room"]; [P2, "Room is comfortable"]) | \\
| Amenities | Two pools ([P1, "There are two pools"]) | \\
| Service | Friendly staff ([P1, "friendly staff"]) \&\&\& overshadowed by unhelpful staff ([P2, "Poor customer service overshadowed the beautiful location"]) | \\
| Food \& Beverage | Exceptional dining experience ([P3, "Exceptional dining"]) \&\&\& limited breakfast options ([P1, "but breakfast was lacking"]) \\
 \\
New Summary Table: \\
| Attribute | Value | \\
| --- | --- | \\
| Food \& Beverage | improved breakfast variety and quality ([P4, "improvements in breakfast variety and quality"])| \\
| Lobby Design | Modern design ([P4, "recently renovated its lobby, which now features a modern design"])| \\
 \\
Combined Summary Table: \\
| Attribute | Value | \\
| --- | --- | \\
| Room Quality | Spacious and comfortable rooms ([P1, "spacious room"]; [P2, "Room is comfortable"]) | \\
| Amenities | Two pools ([P1, "There are two pools"]) | \\
| Food \& Beverage | Exceptional dining experience ([P3, "Exceptional dining"]) \&\&\& limited breakfast options ([P1, "but breakfast was lacking"]) \&\&\& improved breakfast variety and quality ([P4, "improvements in breakfast variety and quality"])| \\
| Lobby Design | Modern design ([P4, "recently renovated its lobby, which now features a modern design"])| \\
 \\
Your Task: \\
Combine existing and new summary tables of the specified entity and generate a new output summary table. \\
Entity: < entity name > \\
 \\
Given Existing Summary Table: \\
< existing summary table > \\
 \\
New Summary Table: \\
< new summary table > \\
 \\
Proceed to combine the two summary tables and generate a new output summary table. Output summary table format should follow the above example of Summary Table. \\
\end{framed}
\caption{Prompt that describes MERGE task along with one example.}
\label{fig:merge-example-prompt}
\end{figure}

\begin{figure}[h]
\begin{framed}
\tt
\small
You will be given two summaries: a reference summary table (gold standard) and a generated summary table. Your task is to check if the gold standard contains the information in the generated summary. \\ 
Please output with Yes/No.\\

Requirements for Yes:\\
 - Meaningful Correspondence: Each attribute-value pair in the generated table should capture the core meaning of its corresponding pair in the reference table, even if worded differently.\\
 - Partially relevant evidence is okay: While the evidence in the generated table does not have to be exactly match with its corresponding attribute-value pair's evidence, it should not be completely off base.
\end{framed}
\caption{LLM-based redundancy checking prompt.}
\label{fig:evidence-finding-prompt}
\end{figure}

%------------------------ Performance across paragraph tones ------------------------
\subsection{Performance across paragraph tones and categories}
\label{app:performance-chart}
\begin{figure*}[t]
    \centering
    \vspace{-30pt}
    \includegraphics[width=1\linewidth]{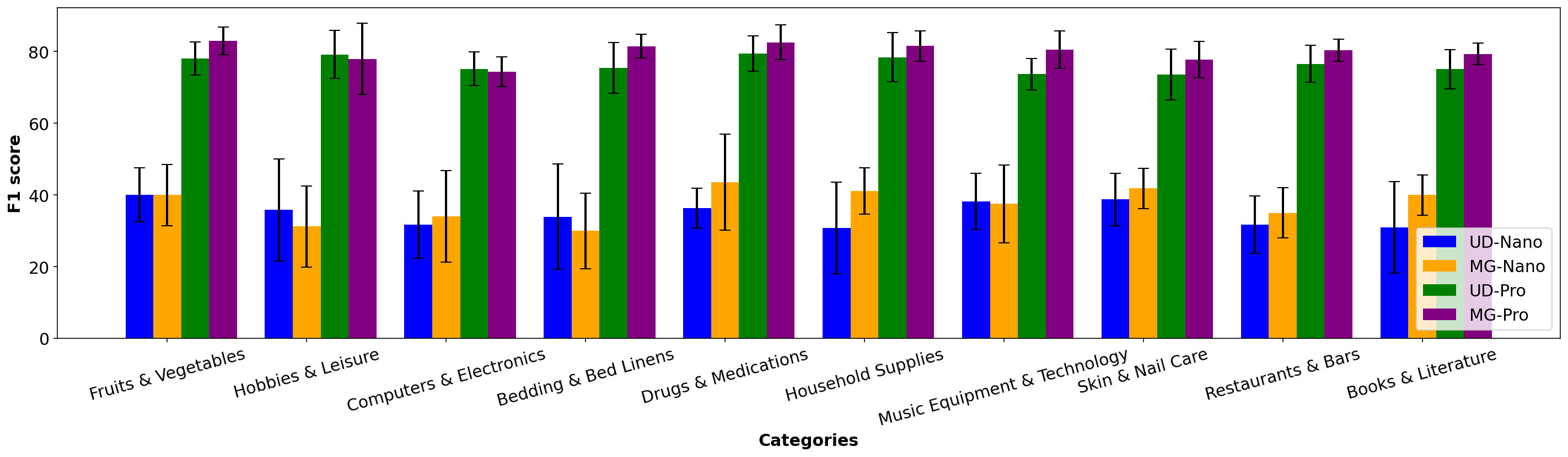}
    \caption{F1 scores across 10 paragraph categories.}
    \label{fig:performance-cate-20}
\end{figure*}

\begin{figure*}[t]
    \centering
    \vspace{-30pt}
    \includegraphics[width=1\linewidth]{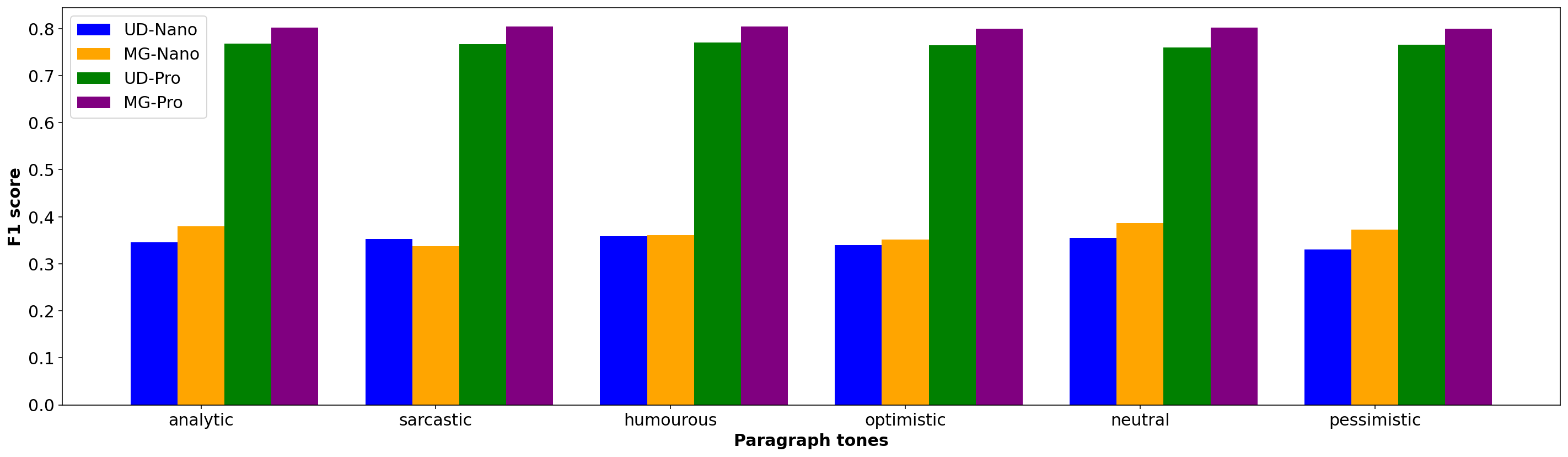}
    \caption{F1 scores across paragraph tones.}
    \label{fig:performance-tone}
\end{figure*}

Figure~\ref{fig:performance-cate-20} shows F1 scores across additional 10 categories. Figure~\ref{fig:performance-tone} presents F1 scores across paragraph tones.

% \begin{figure}[h]
% \begin{framed}
% \tt
% \small
% sample datapoints
% \end{framed}
% \caption{LLM-based redundancy checking prompt.}
% \label{fig:evidence-finding-prompt}
% \end{figure}

\subsection{Example data points of SUMIE dataset}
\label{app:examples}
Figure~\ref{fig:example1}, \ref{fig:example2}, \ref{fig:example3}, \ref{fig:example4} and \ref{fig:example5} show examples of (attribute, value, sentence) triples and distractor sentences exist in our dataset in 5 different categories.

\begin{figure}[h]
\begin{framed}
\tt
\small
\textbf{Category:} Computer \& Video Games \\
 \\
\textbf{Examples of (Attribute, Value, Sentence):} \\
\textbf{Attribute:} Memorability of characters \\
\textbf{Value:} Limited \\
\textbf{Sentence:} GAME1 offers limited memorable characters , making it a forgettable gaming experience. \\
 \\
\textbf{Attribute:} Story \\
\textbf{Value:} Lackluster and predictable storyline \\
\textbf{Sentence:} And don't even get me started on the story  - it's so predictable, I could write it in my sleep! \\
 \\
\textbf{Attribute:} Microtransactions and in-game purchases \\
\textbf{Value:} Optional microtransactions \\
\textbf{Sentence:} The game features optional microtransactions , so you can choose not to spend any additional money. \\
 \\
\textbf{Examples of distractor sentences:} \\
- HUMAN's empathy is a healing potion, allowing them to connect with others and understand their virtual and real-life struggles. \\
- GAME10's characters are complex and relatable, drawing players into the game's world and making them care about the fate of Aloy and her companions. \\

\end{framed}
\caption{Examples of (attribute, value, sentence) triples and distractor sentences in Computer \& Video Games category.}
\label{fig:example1}
\end{figure}

\begin{figure}[h]
\begin{framed}
\tt
\small
\textbf{Category:} Vitamins \& Supplements \\
 \\
\textbf{Examples of (Attribute, Value, Sentence):} \\
\textbf{Attribute:} Brand \\
\textbf{Value:} Longstanding history in the industry \\
\textbf{Sentence:} Vitamin Company1 boasts a long-standing history in the industry , ensuring credibility and trust for their products. \\
 \\
\textbf{Attribute:} Price \\
\textbf{Value:} Not covered by insurance \\
\textbf{Sentence:} But hey, at least it's not covered by insurance . \\
 \\
\textbf{Attribute:} Side Effects \\
\textbf{Value:} May cause mild gas or bloating \\
\textbf{Sentence:} But be warned, this fiber party comes with a side of gas and bloating. \\
 \\
\textbf{Examples of distractor sentences:} \\
- HUMAN's optimism is a probiotic, maintaining a healthy balance in their outlook and promoting a positive gut feeling about the future. \\
- The technology-enabled tracking feature of Vitamin Company10 allows users to monitor their caffeine intake conveniently. \\

\end{framed}
\caption{Examples of (attribute, value, sentence) triples and distractor sentences in Vitamins \& Supplements category.}
\label{fig:example2}
\end{figure}

\begin{figure}[h]
\begin{framed}
\tt
\small
\textbf{Category:} Restaurants \& Bars \\
 \\
\textbf{Examples of (Attribute, Value, Sentence):} \\
\textbf{Attribute:} WiFi Access \\
\textbf{Value:} Convenient for business meetings or working lunches \\
\textbf{Sentence:} This spot offers convenient WiFi access, making it perfect for business meetings or working lunches.  \\
 \\
\textbf{Attribute:} Catering Services \\
\textbf{Value:} Delicious and customizable menus \\
\textbf{Sentence:} And if you're feeling fancy, hit up their catering service. \\
 \\
\textbf{Attribute:} Noise Level \\
\textbf{Value:} Excessively loud and distracting \\
\textbf{Sentence:} Just be warned, it can get loud AF , so if you're trying to have a deep convo, forget about it. \\
 \\
\textbf{Examples of distractor sentences:} \\
- HUMAN's determination is a bustling coffee shop, where the aroma of ambition permeates the air. \\
- RESTAURANT10's edible garden on-site provides fresh, seasonal ingredients that add a touch of vibrancy to their dishes. \\

\end{framed}
\caption{Examples of (attribute, value, sentence) triples and distractor sentences in Restaurants \& Bars category.}
\label{fig:example3}
\end{figure}

\begin{figure}[h]
\begin{framed}
\tt
\small
\textbf{Category:} Books \& Literature \\
 \\
\textbf{Examples of (Attribute, Value, Sentence):} \\
\textbf{Attribute:} Overall Quality \\
\textbf{Value:} A masterpiece of literature \\
\textbf{Sentence:} Step into BOOK1, a literary masterpiece  that will transport you to another realm. \\
 \\
\textbf{Attribute:} Language \\
\textbf{Value:} Written in prose \\
\textbf{Sentence:} AndImmerse yourself in the author's exquisite prose , which paints vivid imagery on the canvas of your mind. \\
 \\
\textbf{Attribute:} Binding \\
\textbf{Value:} Unattractive and unappealing \\
\textbf{Sentence:} While the binding may not be its strong suit , the power of the story within far outweighs its aesthetic shortcomings. \\
 \\
\textbf{Examples of distractor sentences:} \\
- HUMAN's life is a masterpiece, a unique and captivating story that is still being written with every passing day. \\
- BOOK10's books are not only visually stunning but also intellectually stimulating, inviting readers to engage with complex themes and ideas. \\

\end{framed}
\caption{Examples of (attribute, value, sentence) triples and distractor sentences in Books \& Literature category.}
\label{fig:example4}
\end{figure}

\begin{figure}[h]
\begin{framed}
\tt
\small
\textbf{Category:} Education \\
 \\
\textbf{Examples of (Attribute, Value, Sentence):} \\
\textbf{Attribute:} School Culture \\
\textbf{Value:} Lack of support and camaraderie among students \\
\textbf{Sentence:} Welcome to The Evergreen School, where the competition is fierce and the support is scarce . \\
 \\
\textbf{Attribute:} Learning Environment \\
\textbf{Value:} Innovative teaching methods \\
\textbf{Sentence:} But hey, at least they'll be exposed to innovative teaching methods  (if they can keep up with the breakneck pace). \\
 \\
\textbf{Attribute:} Study Abroad Opportunities \\
\textbf{Value:} Immersive experiences in diverse cultures \\
\textbf{Sentence:} If you're looking for immersive experiences in diverse cultures, this school offers study abroad programs  that will expand your horizons. \\
 \\
\textbf{Examples of distractor sentences:} \\
- HUMAN's mind is a fertile ground where ideas bloom and take root, transforming into a thriving garden of understanding. \\
- EDUCATION10's strong industry partnerships provide students with valuable internships and networking opportunities, preparing them for successful careers. \\

\end{framed}
\caption{Examples of (attribute, value, sentence) triples and distractor sentences in Education category.}
\label{fig:example5}
\end{figure}

\end{document}